\documentclass[journal]{IEEEtran}
\usepackage[utf8]{inputenc}
\usepackage[T1]{fontenc}% optional T1 font encoding
\usepackage{graphicx}
\usepackage{times}
\usepackage{helvet}
\usepackage{courier}
\usepackage{amsmath}
\usepackage{algorithm}
\usepackage{algorithmic}
\usepackage{csquotes} 
\usepackage{color}
\usepackage{paralist}
\usepackage{amssymb}
\usepackage{indentfirst}
\usepackage{float}
\usepackage{multirow}
\usepackage{cite}
\usepackage{mathrsfs}

\usepackage{mathrsfs} 
\usepackage{amsfonts}
\usepackage{todonotes}
\usepackage{pgfplots} %引用包
\usepackage{epstopdf}
\usepackage{epsfig}
\pgfplotsset{compat=newest}
\usepackage{color}
\definecolor{forestgreen}{RGB}{0,139,69}

\usepackage{xcolor}
\definecolor{citecolor}{HTML}{0071bc}
\usepackage[colorlinks, linkcolor=red,  anchorcolor=blue, citecolor=citecolor]{hyperref} 
\usepackage{multirow}
\usepackage{xcolor}
\definecolor{SeaGreen4}{RGB}{0,205,102} 
\definecolor{SlateBlue}{RGB}{106,90,205} 
\definecolor{DarkRed}{RGB}{178,34,34} 
	
\usepackage[switch]{lineno}
\usepackage{textcomp,booktabs}
\usepackage{amssymb}% http://ctan.org/pkg/amssymb
\usepackage{pifont}% http://ctan.org/pkg/pifont
\newcommand{\cmark}{\ding{51}}%

\usepackage{makecell}

\usepackage{colortbl}
\definecolor{mygray}{gray}{.9}
\definecolor{mypink}{rgb}{.99,.91,.95}
\definecolor{mycyan}{cmyk}{.3,0,0,0}
\usepackage{graphicx} % 加载图片宏包
\usepackage{subcaption} % 用于子图的排版

\begin{document}

\title{ ReasoningTrack: Chain-of-Thought Reasoning for Long-term Vision-Language Tracking }

\author{Xiao Wang, \emph{Member, IEEE}, Liye Jin, Xufeng Lou, Shiao Wang, Lan Chen*, Bo Jiang*, Zhipeng Zhang   

\thanks{$\bullet$ Xiao Wang, Liye Jin, Xufeng Lou, Shiao Wang, Bo Jiang are with the School of Computer Science and Technology, Anhui University, Hefei, China. 
(email: \{xiaowang, jiangbo, chenlan\}@ahu.edu.cn)}  
\thanks{$\bullet$ Zhipeng Zhang is with School of Artificial Intelligence, Shanghai Jiao Tong University (email: zhipeng.zhang.cv@outlook.com)}
\thanks{$\bullet$ Lan Chen is with the School of Electronic and Information Engineering, Anhui University, Hefei 230601, China. (email: chenlan@ahu.edu.cn)}
\thanks{* Corresponding Author: Lan Chen \& Bo Jiang} 
}

\markboth{IEEE TRANSACTIONS ON ***, 2025}     
{Shell \MakeLowercase{\textit{et al.}}: Bare Demo of IEEEtran.cls for IEEE Journals}

\maketitle

% As a general rule, do not put math, special symbols or citations in the abstract or keywords.
\begin{abstract}
Vision-language tracking has received increasing attention in recent years, as textual information can effectively address the inflexibility and inaccuracy associated with specifying the target object to be tracked. Existing works either directly fuse the fixed language with vision features or simply modify using attention, however, their performance is still limited. Recently, some researchers have explored using text generation to adapt to the variations in the target during tracking, however, these works fail to provide insights into the model's reasoning process and do not fully leverage the advantages of large models, which further limits their overall performance. To address the aforementioned issues, this paper proposes a novel reasoning-based vision-language tracking framework, named ReasoningTrack, based on a pre-trained vision-language model Qwen2.5-VL. Both SFT (Supervised Fine-Tuning) and reinforcement learning GRPO are used for the optimization of reasoning and language generation. We embed the updated language descriptions and feed them into a unified tracking backbone network together with vision features. Then, we adopt a tracking head to predict the specific location of the target object. In addition, we propose a large-scale long-term vision-language tracking benchmark dataset, termed TNLLT, which contains 200 video sequences. 20 baseline visual trackers are re-trained and evaluated on this dataset, which builds a solid foundation for the vision-language visual tracking task. Extensive experiments on multiple vision-language tracking benchmark datasets fully validated the effectiveness of our proposed reasoning-based natural language generation strategy. The source code of this paper will be released on \url{https://github.com/Event-AHU/Open_VLTrack} 
\end{abstract}


\IEEEpeerreviewmaketitle

\section{Introduction}

%% background of vlt 
\IEEEPARstart{V}{isual} Object Tracking (VOT) aims to locate a target object initialized in the first frame across subsequent video frames, producing a bounding box $(x, y, w, h)$ for each frame. This task finds applications in intelligent video surveillance, autonomous driving, and sports analytics. However, due to challenges such as fast motion, objects moving out of view, varying illumination, and adversarial attacks, achieving high-performance visual object tracking remains a significant challenge with much progress yet to be made.

%% the development of vlt 
To address the aforementioned issues, some researchers resort to natural language to improve the tracking performance. Specifically, 
Zhou et al.~\cite{zhou2023joint} propose a unified framework that jointly addresses visual grounding and tracking by modeling relations across multi-source references and test images, 
Ma et al.~\cite{ma2024unifying} propose a unified tracker called UVLTrack, which can flexibly process visual and language input (BBOX, NL, NL+BBOX). Wang et al.~\cite{wang2021towards} propose an adaptive switch based vision-language tracking framework, termed AdaSwitch.  In order to explore the role of text in visual tracking, Guo et al.~\cite{guo2024divert} annotate six popular tracking datasets with general attribute words, providing a strong database for VL tracking and then they propose ModaMixer for unified VL representation learning.

\begin{figure}
\center
\includegraphics[width=0.95\linewidth]{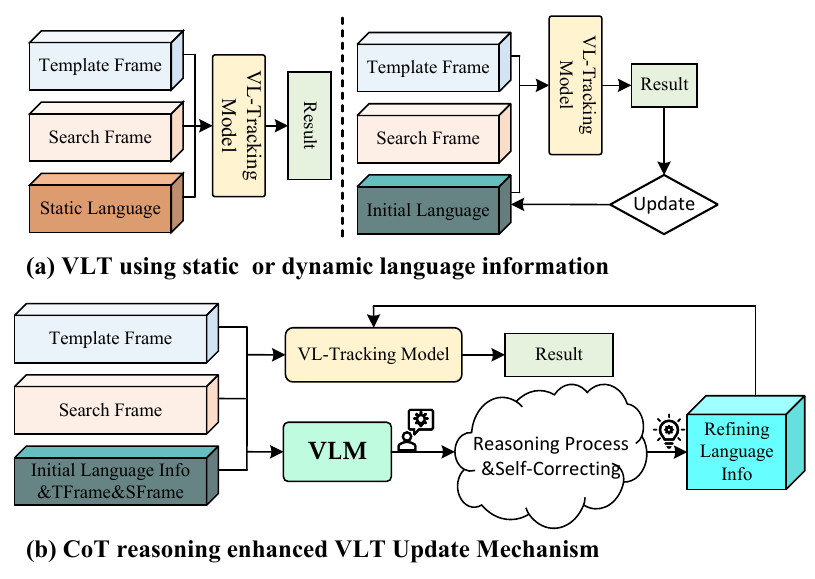}
\caption{Existing vision-language tracking methods either adopt static language information (left) or update language information directly (right), both of which face the problem of insufficient accuracy in language description. We propose a reasoning update framework designed to achieve high accuracy language description.} 
\label{firstIMG}
\end{figure} 

% Early research efforts typically integrated text descriptions directly into visual tracking frameworks, focusing on designing modules for interactive fusion of textual and visual features. However, fixed text descriptions are clearly unable to adapt to changes in the target's appearance. Although some approaches have employed attention mechanisms to weight the text information, their effectiveness has been limited. 

Despite the progress, existing vision-language tracking algorithms are still limited by the following issues: 
1). The natural language specification provided for each video only works for the limited start frames; it fails to adapt to the changing target object. For example, as shown in Fig.~\ref{datasetfig}, the tracked object is no longer suitable for the initial text description as the video length increases. 
2). Current research work takes visual images and initial text descriptions as inputs to dynamically update the natural language specification, thereby improving tracking performance. However, few algorithms can provide a reasonable reasoning process for the text updating procedure, resulting in a lack of interpretability.
3). The visual object tracking community still lacks a large-scale, natural language description-guided visual tracking dataset, especially for long video sequences. This limitation further hinders the practical application of large models in vision-language tracking tasks, preventing full utilization of the benefits brought by the development of large foundation models.  
The above reflections inspire us to consider how to design a vision-language tracking framework with logical reasoning capabilities, which can effectively update the text descriptions dynamically as the tracking progresses, adapting to changes in the target object. More importantly, simply updating text without explicit reasoning may harm trust and reliability in critical applications. Therefore, revealing the rationale behind text adaptations is essential to ensure updates align with actual visual evidence.

% it is crucial to enable users to understand the reasoning process behind these text updates, thereby enhancing the interpretability and credibility of the algorithm.

%% visual reasoning 
Recently, many Large Language Models (LLMs) have demonstrated strong logical reasoning capabilities, such as GPT-4o~\cite{openai2024gpt4ocard}, Deepseek-R1~\cite{deepseekai2025deepseekr1incentivizingreasoningcapability}, and Qwen~\cite{bai2023qwentechnicalreport} series. These models all introduce versions with strong reasoning abilities to address more complex problems. Specifically, Deepseek propose R1 model~\cite{deepseekai2025deepseekr1incentivizingreasoningcapability}, which is trained via large-scale reinforcement learning (RL) without supervised fine-tuning and demonstrates remarkable reasoning capabilities, Qwen3~\cite{yang2025qwen3technicalreport} integrate thinking mode and non-thinking mode into a unified framework, which eliminates the need to switch between different models. Additionally, there has been a growing effort to integrate reasoning into various tasks, such as visual grounding, visual counting, object detection and segmentation. More in detail, Liu et al.~\cite{liu2025seg} propose Seg-Zero designed for reasoning segmentation, and Guan et al.~\cite{guan2025rstar} propose a novel method that could improve the mathematical reasoning ability of LLM. Inspired by the success of reasoning-based models, in this work, we explore the dynamic updating of language specifications using a reasoning-driven approach.

%% our work  
In this work, we propose a novel reasoning based vision-language tracking framework based on Chain-of-Thought (CoT), termed ReasoningTrack. The key insight of this work is to utilize a large Vision-Language Model (VLM) to monitor the tracking process and dynamically update the initialized natural language specification. More in detail, we first embed the given search and template frames into visual tokens and adopt the BERT tokenizer to embed the initial language as the static semantic information, then, feed them into a unified tracking backbone for feature extraction and interactive learning. In addition, we employ the pre-trained vision-language model Qwen2.5-VL to generate a reasoning chain and a dynamic language specification to adapt to the variation of the target object. The SFT (Supervised Fine-Tuning) and reinforcement learning GRPO are used for the optimization of reasoning and language generation. The updated language description is embedded and fed into the unified backbone network. Then, we adopt a tracking head to predict the specific location of the target object. An overview of our proposed ReasoningTrack framework can be found in Fig.~\ref{framework}.

To address the aforementioned third issue, this paper proposes a new long-term vision-language tracking dataset called TNLLT. The raw videos for this dataset were collected from video websites, primarily involving TV, movies, games, entertainment, documentaries, and more. We performed meticulous rectangular box annotations on the cropped videos and labeled language descriptions of object appearance, motion, and other cues. These videos encompass 15 challenging factors related to vision-language tracking. Statistically, the dataset includes 200 videos, each at 30 FPS, with an average length of 2,729 frames, and the maximum length reaching up to 14,976 frames. We benchmark 20 representative and state-of-the-art visual trackers involving different initializations (i.e., bounding box (BBox) only, language only, and BBox + language) for future works to compare. We believe that the construction of this benchmark can lay a solid foundation for subsequent research by providing a high-quality dataset and a basis for benchmark algorithm comparisons.

%% contributions 
To sum up, the main contributions of this paper can be summarized as the following three aspects: 

1). We propose a novel natural language reasoning strategy for vision-language tracking based on pre-trained large vision-language models, termed ReasoningTrack. It monitors the tracking procedure and updates the natural language descriptions with a reasoning process, which makes the tracking more robust and intelligent. 

2). We propose a large-scale long-term vision-language tracking benchmark dataset, termed TNLLT. It contains 200 video sequences, 150/50 for training and testing, respectively. 20 baseline visual trackers are re-trained and evaluated on this dataset, which builds a solid foundation for the vision-language visual tracking task. 

3). Extensive experiments on three widely used tracking benchmark datasets (i.e., OTB-Lang~\cite{wu2013online}, GOT-10K~\cite{huang2019got}, TNL2K~\cite{wang2021tnl2k}) and the proposed TNLLT dataset fully validated the effectiveness of our proposed vision-language tracking framework.

\textbf{The following of this paper is organized as follows:} 
In Section~\ref{sec::relatedWorks}, we introduce the related works most relevant to this paper from the visual tracking, visual-language tracking, and chain-of-thought reasoning. 
In Section~\ref{sec::Methodology}, we first give an overview of our framework, then we focus on the input representation, the preliminary baseline tracking framework, and the chain-of-thought reasoning module. After that, we introduce the training and testing details for our tracking framework. 
In Section~\ref{sec::benchmarkdataset}, we mainly discuss the large-scale benchmark dataset TNLLT, with focus on the data collection and annotation, attribute definition, and statistical analysis. 
In Section~\ref{sec::experiments}, we dive into the details of the dataset and evaluation metric, implementation details, ablation study, comparison with other state-of-the-art (SOTA) trackers on public benchmark datasets, and also the visualizations. 
Finally, we conclude this paper and point out the research directions for the vision-language tracking in Section~\ref{sec::Conclusion}.

% \begin{figure*}[!htp]
% \center
% \includegraphics[width=7in]{figures/frontIMG.jpg}
% \caption{sss.} 
% \label{frontIMG}
% \end{figure*}

\section{Related Work} \label{sec::relatedWorks}

In this section, we review the related works with a focus on visual object tracking~\footnote{\url{https://github.com/wangxiao5791509/Single_Object_Tracking_Paper_List}}, vision-language tracking, and chain-of-thought reasoning.

\subsection{Visual Tracking}   
In computer vision, visual-based tracking is an extremely challenging task~\cite{9339950}. Specifically, Single Object Tracking (SOT) aims to detect positions of the target object in each subsequent frame based on the first frame. Usually, RGB serves as the primary tracking modality, with methods primarily including binary classification based~\cite{hare2016struck,Nam2015Learning,SongYiBing_2018_CVPR,dai2020LTMU}, Siamese network based~\cite{bertinetto2016fully,Tao2016Siamese,li2018siamRPN,wang2019SiamMask,xu2020siamfc++,mayer2022ToMP,chen2021TransT}, correlation filter based~\cite{danelljan2015learning, choi2017attentional, Bolme2010Visual, yao2018joint, valmadre2017end} and discriminative learning based~\cite{danelljan2019atom, bhat2019dimp, danelljan2020prdimp, bhat2020KYS}. Inspired by the Transformer architecture~\cite{2017Attention}, the majority of current research is based on this framework. For example, Ye et al.~\cite{ye2022ostrack} propose a simple one-stream tracking framework based on Vision Transfromer. Chen et al.~\cite{sutrack} propose SUTrack, which consolidates five SOT tasks into a unified model. Gao et al.~\cite{gao2022aiatrack} present an attention in attention module to improve the attention mechanism for Transformer visual tracking. Recent studies have argued that static reference templates in visual tracking can limit accuracy, particularly in scenarios involving significant appearance variations, occlusion, or deformation. Li et al.~\cite{li2025dynamic} propose DUTrack which enhances tracking capability by dynamically updating multi-modal references. To further advance the dynamic updating paradigm in visual tracking, our work explores the potential of leveraging large pre-trained models to enhance reference information updating during the tracking process. 

% Trackers can be categorized into two groups depending on whether the reference information is dynamically updated throughout the tracking process. The first category comprises static reference trackers, which maintain a fixed set of reference information once initialized and do not alter it during tracking such as ***. The second category includes dynamic reference trackers, which continuously update the reference information based on new observations during the tracking process. This adaptability allows them to better handle variations in the target's appearance. For example, ***. 

\subsection{Visual-Language Tracking}   
In recent years, with the rise of large language models, an increasing number of researchers have been focusing on integrating semantic information into the tracking process. Unlike traditional tracking methods, Visual-Language Tracking(VLT) methods provide both language description and bounding box of the target object in the initial frame, which could handle the appearance variation of target object as a high-level sematic information. The origins of Visual-Language tracking can be traced back to Li's work~\cite{li2017tracking}, where they providing one sentence for each
video and use LSTM~\cite{graves2012long} and VGG~\cite{simonyan2014very} to extract the features of given text and video frame. After that, Wang et al.~\cite{wang2018describe} propose DAT to track the target object based on the provided BBox and its language information. Yang et al.~\cite{yang2020grounding} proposed a new GTI framework for VLTracking, which decompose the task into three sub tasks: grounding, tracking, and integration. Wang et al.~\cite{wang2021towards} propose a standard benchmark named TNL2K and introduce a method based on an adaptive local-global-search scheme to achieve precise vision-language tracking. Guo et al.~\cite{guo2022divert} propose a unified-adaptive VL representation to achieve SOTA tracking without complex Transformer. Zhou et al.~\cite{zhou2023joint} propose JointNLT to connect grounding and tracking task in a joint framework. All-in-one~\cite{zhang2023all} has desinged a new framework for multi-modal VL Tracking. MMTrack~\cite{zheng2023toward} reformulates visual-language tracking as a token generation task and proposes a novel pipeline, unleashing the potential of multimodal learning through unified modeling. UVLTrack~\cite{ma2024unifying} designs a modality-unified feature extractor for joint visual and language feature learning. With the development of large language models, many researchers have begun exploring how to leverage LLM's capabilities to achieve more accurate and robust visual-language tracking. ChatTracker~\cite{sun2024chattracker} utilizes the Multimodal Large Language Model to enhance the performance of visual tracking. DUTrack~\cite{li2025dynamic} use BLIP~\cite{li2022blip} to generate dynamic language description, addressing the challenge of static textual queries in vision-language tracking. Different from these works, we aim to achieve more robust tracking results for longer video sequences by dynamically updating the initial linguistic descriptions of the tracker.

\subsection{Chain-of-Thought Reasoning} 
Recent years have witnessed remarkable progress in Large Language Reasoning Models (LLRMs) such as DeepSeekR1~\cite{deepseekai2025deepseekr1incentivizingreasoningcapability}, Gemini2 Flash Thinking and Claude3.7 Sonnet. These models will provide a reasoning process before giving an answer. The earliest chain of thought methods can be traced back to Wei's work~\cite{wei2022chain}, which propose a few shot CoT method to enhance model's reasoning ability. After that, Kojima T et al.~\cite{kojima2022large} propose a zero shot method to create a thinking chain model which provides an additional sentence `Let's think step by step'. Another method to enable the model to have reasoning capabilities is to directly provide the model's thinking process in the prompt, such as Zhang's work~\cite{zhang2024supervised}. Unlike the work mentioned above, more and more researchers are beginning to focus on fine-tuning large language models to enhance model's reasoning abilities. A promising direction involves fine-tuning LLMs with domain-specific knowledge to improve the model's understanding of complex tracking scenarios. Furthermore, techniques such as Reinforcement Learning from Human Feedback (RLHF) have been successfully applied to align model outputs with human preferences~\cite{ouyang2022training}. For example, DeepSeek-R1~\cite{deepseekai2025deepseekr1incentivizingreasoningcapability} independently stimulates the reasoning capabilities of models through pure reinforcement learning (RL). R1-VL~\cite{zhang2025r1} introduces StepGRPO, a novel approach that enables multimodal large language models (MLLMs) to autonomously enhance their reasoning capabilities through an efficient reward-based self-improvement mechanism. In terms of downstream tasks, Seg-Zero~\cite{liu2025seg} introduces an innovative approach that successfully integrates Chain-of-Thought (CoT) reasoning into segmentation tasks through pure reinforcement learning. R1-Track~\cite{wang2025r1} proposes a novel framework that combines SFT and RL to effectively transfer multimodal large language models (MMLMs) capabilities to visual tracking tasks. In this work, we did not directly obtain the tracking results of the target using MLLM, instead, we utilized the high-quality linguistic modal information it provided to enhance the tracking process.

\section{Methodology}  \label{sec::Methodology}

\begin{figure*}[!htp]
\center
\includegraphics[width=6.5in]{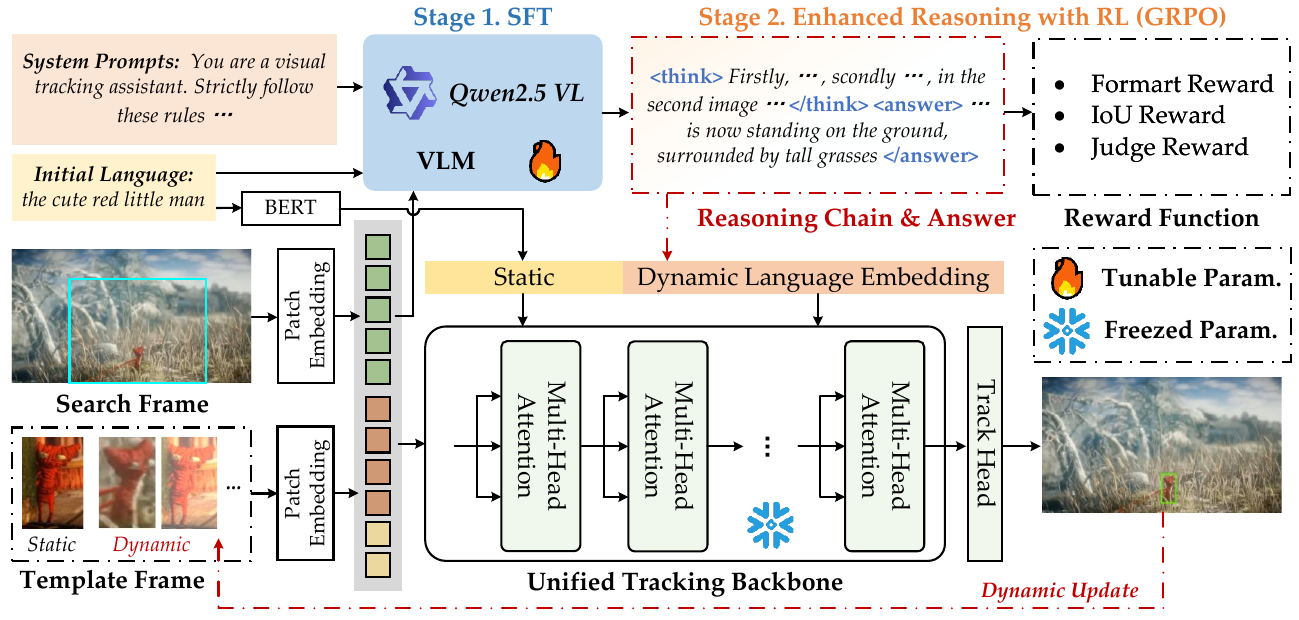}
\caption{\textbf{An overview of our proposed ReasoningTrack Framework for Long-term Vision-Language Tracking.} It contains a Multimodel Large Language Model and a Unified Tracking Backbone. Template frame, search frame and Optimized language description are fed into the unified tracking backbone to get the tracking result. Specifically, we first train the Vision Language Model on a supervised fine-tuning datasets, enhancing the model's instruction-following ability and Chain of Though ability. Subsequently, by employing reinforcement learning strategies, we not only improve the quality of the output text from the large language model but also enhance the performance of the tracking model. The method can be seen as a plug-and-play module that can seamlessly integrate into other visual language tracking frameworks to improve tracking performance.}
\label{framework}
\end{figure*}

In this section, we will provide a detailed introduction to the proposed ReasoningTrack. First, we briefly introduced the tracking task and the overall framework of ReasoningTrack. After that, we introduce the method for constructing the Reasoning dataset and the input representation. Finally, we detail how to fine-tune multimodal large language models using SFT and RL.

\subsection{Preliminaries}  
\textbf{Task Formulation}
Give a video sequence $V = \{s_i\}_{i=1}^{i=T}$, a bounding box $bbox = [x,y,w,h]$ and a language description $L=\{w_0,w1,...,w_n\}$, where $T$ represents the total length of the video sequence, $n$ represents the length of the language. Visual language tracking aims to obtain the tracking results $R=\{[x_i,y_i,w_i,h_i]\}_{i=1}^{i=T}$ for each frame using the information provided above.

\textbf{Input Representation}
The input of the unified tracking backbone consists of three components: the search frame $S\in R^{3\times H_{s} \times W_{s}}$, the template frame $T\in R^{3\times H_{t} \times W_{t}}$ and the language information $L$, where $H$ and $W$ represent the size of the search area and the template area. Then, $S$ and $T$ are converted into token sequences via image patch embedding, and $L$ is also tokenized into a sequence of embeddings via BERT's~\cite{devlin2019bert} tokenization process. During the training phase of the large model, our inputs also consist of three parts: search frames, template frames, and language information. Unlike the tracking network, this stage does not crop the original frames into small regions. Instead, both the search and template images are fed into the model at full size.
% The input of our method consists of two components: images $I\in R^{3\times H \times W}$ and text $L$, where $H$ and $W$ represent the height and weight of the image. After receiving the input data, $I$ are mapped to tokenized representations through patch embedding operation and $L$ is also tokenized into a sequence of embeddings via BERT's~\cite{devlin2019bert} tokenization process. Next, we will input the integrated multimodal information into a Unified Tracking Backbone for modality fusion, and ultimately delivered to a Track Head to obtain the tracking results. In order to make this model better adapt to the appearance variability of objects, we introduce dynamic template updates. While this can address the issue of dynamic changes in the target object to some extent, static text may mislead the subsequent tracking process. Therefore, we incorporate a large language model to accomplish dynamic text updates.

\subsection{Overview} 
\begin{figure*}[!htp]
\centering
\includegraphics[width=\linewidth]{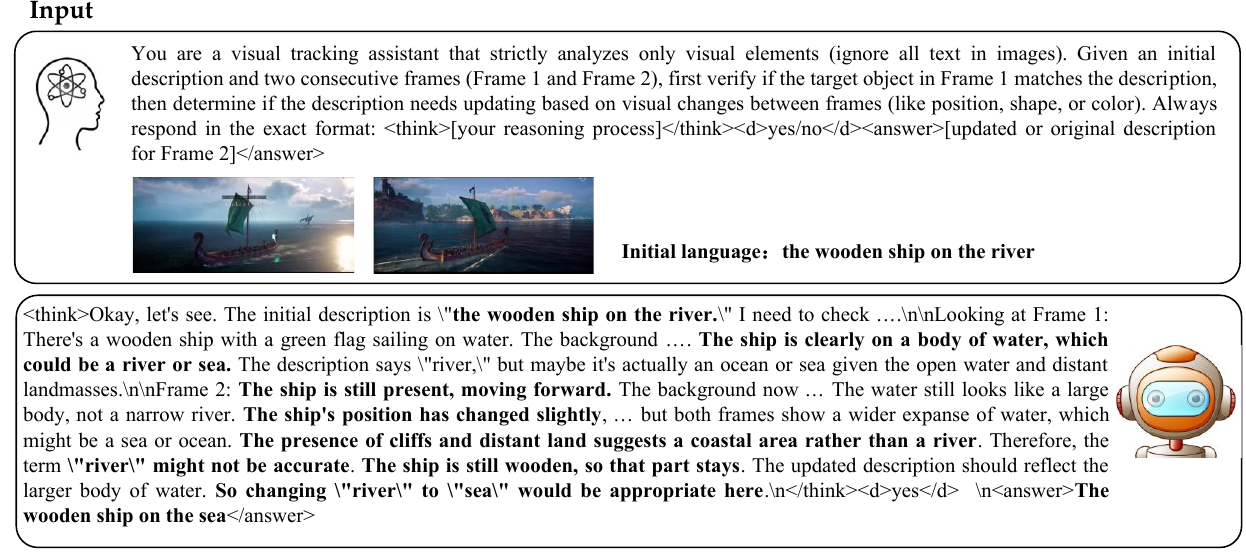}
\caption{Prompt used for generating Chain-of-Thought dataset.}
\label{anno}
\end{figure*}

An overview of ReasoningTrack is shown in Fig.~\ref{framework}.
Our framework consists of two main components: a unified tracking backbone and a simple pipeline for fine-tuning the Large Language Model. The input of our ReasoningTrack consists of three parts: the search frame, the template frame, and the language information. 
% After that, the input enters the multimodal large language model (Qwen2.5 VL) to get a refined language description. Then the refined language description, search frame, and template frame are sent into a frozen unified tracking backbone and tracking head to obtain the tracking results. Finally, this result is used to calculate the Intersection over Union (IoU) reward in reinforcement learning and optimize Qwen2.5 VL.
These inputs first pass through the multimodal large language model (Qwen2.5-VL) to generate enhanced textual descriptions. The refined description, along with the visual frames, is then fed into the frozen unified tracking architecture to produce tracking predictions. Finally, we compute Intersection-over-Union (IoU) scores from these predictions to serve as reinforcement learning rewards for optimizing the Qwen2.5-VL model.

\subsection{Reasoning Data Generation and SFT based Optimization}  
Recent studies~\cite{deepseekai2025deepseekr1incentivizingreasoningcapability} have shown that supervised fine-tuning on longer chain-of-thought datasets can effectively improve the model's reasoning ability. In order to lay a solid foundation for the fine-tuning of the reinforcement learning stage, we first use the Chain-of-Thought dataset to supervise fine-tuning large language model. Specifically, we first randomly sampled 197, 200, 200, 200, and 200 template-search frame pairs from the GOT10k~\cite{huang2019got}, LaSOT~\cite{fan2019lasot}, OTB99~\cite{li2017tracking}, TNL2k~\cite{wang2021tnl2k} datasets, and our proposed TNLLT dataset. In this step, to ensure the collection of high-quality sample pairs, we excluded pairs that were completely obstructed or removed from the field of view, the final remaining dataset is denoted as $D_{sample}$. After that, we input $D_{sample}$ into the Qwen-QVQ-Max model which supports visual input and logical chain output to generate reasoning data as Fig.~\ref{anno}. And then we construct a reasoning dataset denoted as $D_{SFTreason}=(x^i_{S},x^i_{T},lan,Y^i)^{997}_{i=1}$, where $x_S$ denotes  the search frame, $x_T$ denotes the template frame, $lan$ denotes the initial language description and $Y$ denotes the reasoning data generated from Qwen-QVQ-Max. Finally, we employ supervised fine-tuning on the chain-of-thought dataset to cold-start the base MLLM, endowing the model with fundamental reasoning capabilities.

% \subsection{Input Representation}  

% \subsection{SFT based Optimization} 

\subsection{GRPO based Optimization} 
Although the model demonstrates initial reasoning capabilities after supervised fine-tuning based on Chain-of-Thought (CoT) data, its performance still exhibits significant limitations. To further enhance the model's logical reasoning efficiency, we implement reinforcement learning on the basis of the preliminary supervised fine-tuned model. Considering the high computational costs associated with traditional reinforcement learning methods such as PPO, this study employs the GRPO method for model fine-tuning, aiming to improve computational efficiency while ensuring training stability. Specifically, we first sampled nearly 5k RL training samples from GOT10k~\cite{huang2019got}, LaSOT~\cite{fan2019lasot}, OTB99~\cite{li2017tracking}, TNL2k~\cite{wang2021tnl2k} and our proposed TNLLT dataset. Similar to the supervised fine-tuning process, we eliminated unreliable samples, ultimately forming dataset $D_{RL} = (x^i_{S},x^i_{T},lan,box1,box2)^{4928}_{i=1}$, where $x_{S}$ and $x_{T}$ denote template frame and search frame respectively, $lan$ denotes the initial language description and $box1, box2$ denote the object's ground truth in search frame and template frame respectively.

Reward functions play an important role in reinforcement learning because they serve as the primary mechanism for guiding an agent's behavior and shaping its learning process. We manually design the following four reward functions for reinforcement learning.

\textbf{Format Reward Function} This reward is designed to force the model output the specified structure format. Specifically, it guides the model to generate text output that conforms to the $<think></think><d></d><answer></answer>$ format, where the model will put the thinking process in $<think>$ tag, put the judgment of whether to update the text within the $<d>$ tag and the updated text within the $<answer>$ tag. We have designed two formatting rewards to guide the model in progressively learning better output formats. The first format reward only requires the model's response to contain three identifiers: $<think>$, $<d>$, and $<answer>$ and the second format reward necessitates the correct order of the three markers and that $<d>$ contains only `yes' or `no'.

\textbf{IoU Reward Function} The IoU rewards are used to provide effective supervision for the accuracy of language generation by the model. Specifically, we first obtain the text description optimized by MLLM, and then send it along with the template frame and search frame into a frozen tracking backbone to obtain the tracking results, denotes as $box_{pre}$. Finally, the IoU Reward Function is defined as:
\begin{equation}
\mathrm{IoUReward} = 
\begin{cases} 
\mathrm{IoU}(box_2, box_{\mathrm{pre}}) & \text{if }\; \mathrm{IoU} > \theta, \\
0 & \text{otherwise,}
\end{cases}
\tag{1}
\end{equation}
where the IoU function is used to calculate the intersection and union ratio of the two boxes and $\theta$ serves as the control threshold for IoU rewards. In this work, we set $\theta=0.61$ as the threshold, motivated by our empirical observation that the mean IoU during testing approximates this value. This configuration optimally incentivizes the model to achieve rewards.

\textbf{Judge Reward Function} The motivation behind this reward function is to enable the MLLM to autonomously determine the timing of language updates. Our idea is that if the tracking results obtained from the updated text produced by the model are not as good as those obtained from the original text, the model should output a signal to reject the update; otherwise, it should output a signal to update the text. Specifically, we first calculate the IoU of the output target box and the ground truth under the initial text and optimized text settings, denotes as $iou_{1}$ and $iou_2$. Then the judge reward function is defined as:
\begin{equation}
\text{JudgeReward} = 
\begin{cases} 
1 & \begin{aligned}
     &\text{if } (d=\text{yes} \text{ and } \text{iou}_1 < \text{iou}_2) \\
     &\quad \text{or } (d=\text{no} \text{ and } \text{iou}_1 \geq \text{iou}_2),
   \end{aligned} \\
0 & \text{otherwise,}
\end{cases}
\end{equation}
where $d$ is the content extracted from the tag $<d>$ in the MLLM's reply.

% \subsection{Tracking Head and Loss Function}  
% During the testing phase, we replaced the text update module of our baseline DUTrack while maintaining its original multi-modal interaction capabilities and template frame updating abilities. 
The final reward function is defined as:
\begin{equation}
\begin{split}
\text{Overall}_{\text{reward}} = {} & w_{1,2} \times \text{FormatReward}_{1,2} \\ 
                                   & + w_3 \times \text{IoUReward} + w_4 \times \text{JudgeReward},
\end{split}
\end{equation}
where the $w_{i}$ denotes the weights of the four rewards. Finally, we employ the GRPO algorithm to optimize this reward. Specifically, we conduct five samples per question to obtain different responses from the large model, denoted as $r_i,\ 0\leq i \leq 5$, and then calculate their rewards separately. After that, GRPO evaluates the relative quality of responses by normalizing their mean and standard deviation:
\begin{equation}
A_i = \frac{r_i - \mu}{\sqrt{\frac{1}{n} \sum_{i=1}^n (x_i - \mu)^2}},
\end{equation}
where $\mu=(\sum_{i=1}^{n}r_i)/n$, $A_i$ represents the relative quality of the i-th answer. GRPO enables the model to align more closely with answers that have higher advantages. The final training objective incorporates a KL-divergence term $D_{KL}$ to ensure that the optimized policy $\pi_{\theta}$ does not deviate significantly from the parameters $\pi_{base}$ of the original base MLLM. The training objectives of reinforcement learning are as follows:
\begin{equation}
\label{eq:grpo_objective}
\begin{split}
\mathcal{J}_{\text{GRPO}}(\theta) = \mathbb{E}_{(s_t,a_t)\sim\mathcal{D}}
\bigg[&\frac{\pi_\theta(a_t|s_t)}{\pi_{\text{old}}(a_t|s_t)}
\hat{A}_t(s_t,a_t) \\
&- \beta \mathcal{D}_{\text{KL}}\big(\pi_\theta(\cdot|s_t) \parallel \pi_{\text{base}}(\cdot|s_t)\big)
\bigg],
\end{split}
\end{equation}
where $s_t$ denotes the input context of the large language model at time step t, $a_t$ is the next token selected from the vocabulary by the model based on its current state and $\beta$ denotes a regularization coefficient that constrains excessive deviation from the reference policy during optimization.

\begin{algorithm}
\caption{Algorithms for visual language tracking in the testing phase}
\label{alg:llm_tracking}
\raggedright
\textbf{Input}: Input frames $I$, pre-trained tracker $B$, initial language description $L$, fine-tuned multimodal large language model $MLLM$, system prompts $P_{sys}$ \\
\textbf{Parameter}: number of frames $N$, update interval $u$ \\
\textbf{Output}: The tracking results $O$
\vspace{-5mm}
\begin{algorithmic}[1]
    \STATE Initialize the tracker $B$ 
    \STATE Initialize the dynamic language $L_D = L$, the last updated frame number $N_{pre} = 1$
    \FOR{$t = 1$ to $N$}
        \STATE Get dynamic template frames $T_{1,2,3}$
        \STATE Get the tracking result for the current frame $O_t = B(T_{1,2,3},S_t,L_D)$
        \IF{$t$ mod $u$ }
            \STATE $L_D = MLLM(I_{N_{pre}},I_{t},L,P_{sys})$
            \STATE $N_{pre} = t$
        \ENDIF
    \ENDFOR
    \STATE Return $O$
\end{algorithmic}
\end{algorithm}

\begin{table*}
\center
% \scriptsize 
\caption{Comparison of current datasets for object tracking. $\#$ denotes the number of the corresponding item. Lang-A and Lang-I denote that the dataset can be used for language-assisted and initialized tracking tasks. SAV denotes that the dataset contains many videos with significant appearance variation. Adv means the dataset contains adversarial samples (i.e., malicious attacks). DA is short for domain adaptation. RD denotes that the dataset contains reasoning data. VFA means the dataset contains Variable frequency audio information.} \label{benchmarkList}
\resizebox{\textwidth}{!}{
\begin{tabular}{l|ccccccccccccccccc}
\hline \toprule [0.5 pt]
\textbf{Datasets}    &\textbf{\#Videos}  &\textbf{\#Min} &\textbf{\#Mean} &\textbf{\#Max} &\textbf{\#Total} &\textbf{\#FR}  &\textbf{\#Att.} &\textbf{Aim} &\textbf{Absent} &\textbf{Lang-A}  &\textbf{Lang-I} & \textbf{SAV}      &\textbf{Adv}   &\textbf{DA} &\textbf{RD} &\textbf{VFA} \\ 
\hline
\textbf{OTB50 \cite{wu2013online}}   	    &51       &71 	&578    &3,872    &29K    &30 fps      &11    &Eval      &       &     &           &             &          &           \\
\textbf{OTB100 \cite{wu2013online}}   	&100     &71  	&590    &3,872    &59K    &30 fps     &11    &Eval    &      &         &           &          &         &            \\
\textbf{TC-128 \cite{Liang2015Encoding}}   	    &128     &71  	&429    &3,872    &55K    &30 fps       &11    &Eval     &    &      &            &         &            &             \\
\textbf{VOT-2017 \cite{kristan2016novel}}   &60      &41  	&356    &1,500    &21K    &30 fps     &-    &Eval    &       &     &     &                 &             &          \\
\textbf{NUS-PRO \cite{li2015nus}}   	&365     &146  	&371    &5040    &135K    &30 fps      &-    &Eval    &       &      &          &            &          &            \\
\textbf{UAV123 \cite{benchmark2016benchmark}}   	&123     &109  	&915    &3085    &113K    &30 fps         &12    &Eval    &       &      &     &            &               \\
\textbf{UAV20L \cite{benchmark2016benchmark}}   	&20     &1717  	&2934    &5527    &59K    &30 fps          &12    		&Eval    &       &       &  &           &              &                \\
\textbf{NfS \cite{kiani2017need}}   	            &100     &169  	&3830    &20665    &383K    &240 fps           &9    &Eval    &       &         &      &            &          \\
\hline
\textbf{TrackingNet \cite{muller2018trackingnet}} 	&30,643     & - 		&480   	 & -   		& 14.43M   	&30 fps           &15    &Train/Eval    &         &     &       &           &           &              \\
\textbf{OxUvA \cite{valmadre2018long}}   	    		&366     	  & 900 	&4260   &37440   &1.55M  		&30 fps           &6    &Train/Eval     &         &    &         &            &           &           \\
\textbf{GOT-10k \cite{huang2019got}}   			&10,000    &29  		&149    &1,418     &1.5M    		&10 fps    		&6    &Train/Eval   &\cmark    &    &       &             &           &           \\ 
\textbf{LaSOT \cite{fan2019lasot}}   	    				&1,400      &1000		&2506   &11397    &3.52M      &30 fps         	&14    &Train/Eval  &\cmark           &\cmark    &          &       &        &            \\
\textbf{TNL2K~\cite{wang2021tnl2k}}  &2,000     &21  	&622    &18488    &1.24M    &30 fps       &17   &Train/Eval          &\cmark   &\cmark    &\cmark           &\cmark      &\cmark         &\cmark          \\
\textbf{WebUAV-3M~\cite{10004511}}  & 4500 &  40	& 710 &   18841 &    3.3M   & 30fps   &  -  & Train/Eval  &  \cmark  &    \cmark    &    \cmark  &    \cmark     &      &    \\
\textbf{WebUOT-1M~\cite{zhang2024webuot}}  & 1500 &  49	&  733  &  9985  &  1.1M  & 30fps &-  &  Train/Test &  \cmark &\cmark       & \cmark  &\cmark  &  \cmark  &           &      &    \\
\hline
\textbf{TNLLT (Ours)}  & 200  &  1002	&  2729  &  14,976  &  0.55M  &   30 fps    &15 & Train/Eval & \cmark  &  \cmark  & \cmark  &  \cmark & \cmark  &  \cmark &  \cmark &  \cmark \\
\hline \toprule [0.5 pt]
\end{tabular}
}
\end{table*}

\section{TNLLT Dataset} \label{sec::benchmarkdataset}

\subsection{Data Collection and Annotation}
\begin{figure*}[!htp]
\centering
\includegraphics[width=\linewidth]{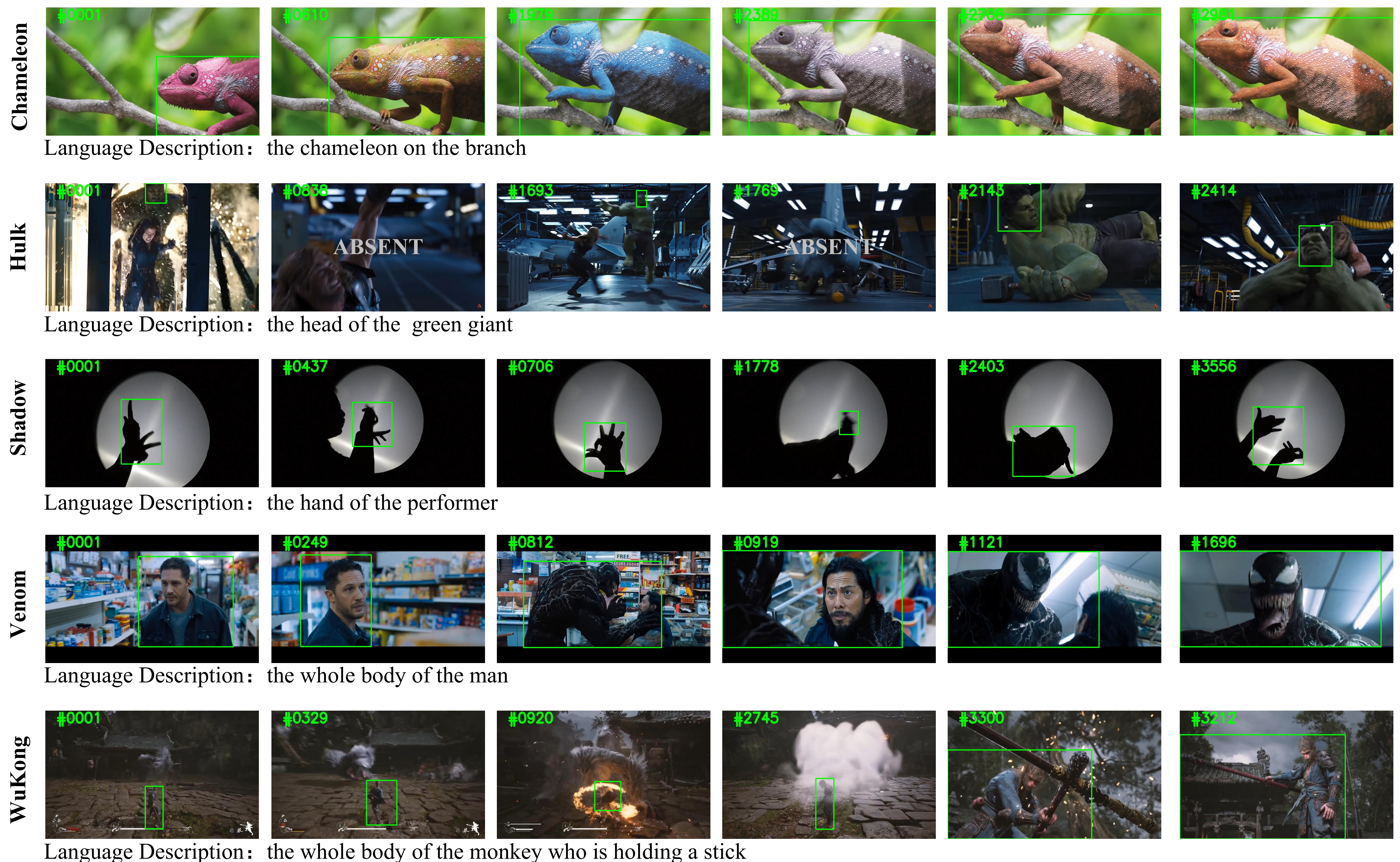}
\caption{Distribution visualization of challenging factors, category of the target object, and bounding box.}
\label{datasetfig}
\end{figure*}

The proposed TNLLT dataset contains 200 video sequences, and most of them are collected from video websites, primarily involving television, movies, games, entertainment, and documentaries. It is noteworthy that our dataset also includes a meticulously sampled reasoning chain dataset, which could provide valuable resources for subsequent researchers to explore more interpretable and robust AI models. We have provided detailed rectangular frame annotations on the edited video, marking language information, audio information, and the attributes of each segment of the video. Specifically, for each video, we annotate one English sentence and delineate one bounding box per frame. Subsequently, we utilize the pyttsx3 library to generate audio information in six distinct speeds based on the language annotations. The left corner point \( (x, y) \), width \(w\) and height \(h\) of the target's bounding box are used as the ground truth, i.e., \([x,y,w,h]\). The annotated description in the first frame indicates the target object's spatial position, relative location, attributes, category, and properties. We also annotate the absent label to enrich the dataset with explicit information regarding the absence of specific objects. Example sequences and annotations are illustrated in Fig.~\ref{datasetfig}. 

\begin{table}
\center
\small  
\caption{Description of 15 attributes in our TNLLT dataset.} 
\label{AttributeList}
\resizebox{0.5\textwidth}{!}{ 
\begin{tabular}{l|lcccccccccccccc}
\hline \toprule [0.5 pt]
\textbf{Attributes}    &\textbf{Description}  \\ 
\hline
\textbf{01. CM}   	    	&Abrupt motion of the camera \\	
\textbf{02. ROT}   	    &Target object rotates in the video \\	
\textbf{03. DEF}   	    &The target is deformable \\
\textbf{04. FOC}   	    &Target is fully occluded \\
\textbf{05. IV}   	    &Illumination variation \\	
\textbf{06. OV}   	    	&The target completely leaves the video sequence \\ 
\textbf{07. POC}   	    	&Partially occluded \\ 
\textbf{08. VC}   	    	&Viewpoint change  \\
\textbf{09. SV}   	    	&Scale variation  \\
\textbf{10. BC}   	    	&Background clutter  \\
\textbf{11. MB}   	    &Motion blur  \\
\textbf{12. ARC}         &The bounding box aspect ratio is outside the range [0.5, 2] \\ 
\textbf{13. LR}         &Low resolution \\ 
\textbf{14. FM}          &Fast motion \\ 
\textbf{15. AS}          &Influence of adversarial samples \\ 
\hline \toprule [0.5 pt]
\end{tabular}
} 
\end{table}

\subsection{Attribute Definition}
\begin{figure*}[!htp]
\center
\includegraphics[width=6.6in]{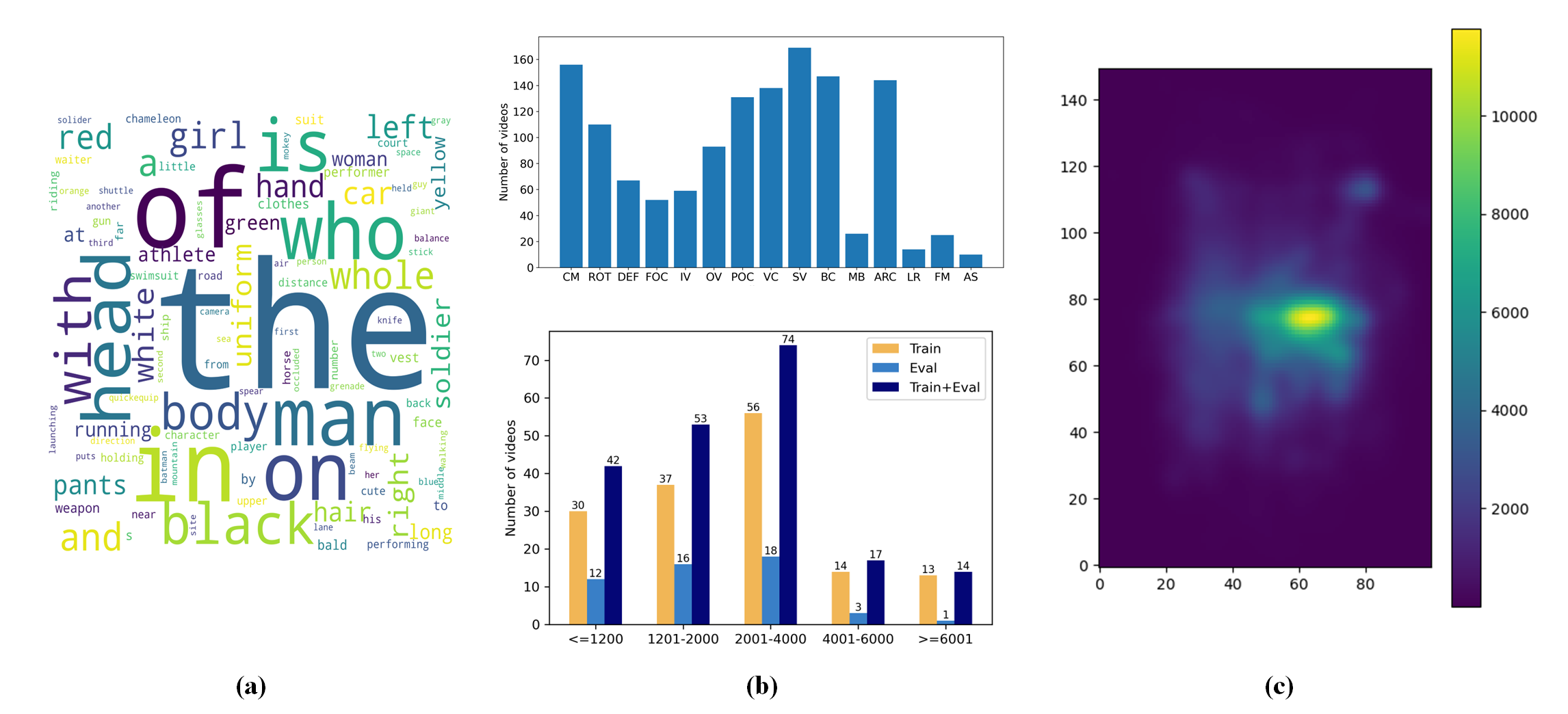}
\caption{(a) Some words in our language description; (b) Distribution of sequences in each attribute and length in our TNLLT. Best viewed by zooming in.; (c) Distribution of bounding box}
\label{datasetInfo}
\end{figure*}

Following popular tracking benchmarks\cite{fan2019lasot,huang2019got,wu2013online,wang2021towards}, we define 15 attributes for each video sequence to evaluate performance under various challenging factors. As shown in Table~\ref{AttributeList}, our proposed TNLLT dataset has the following 15 attributes: CM(Camera Motion), ROT(Rotate of Target), DEF(Deformation), FOC(Fully Occluded), IV(Illumination Variation), OV(Out of View), POC(Partially Occluded), VC(Viewpoint Change), SV(Scale Variation), BC(Background Clutter), MB(Motion Blur), ARC(Aspect Ratio Change), LR(Low Resolution), FM(Fast Motion) and AS(Adversarial Sample). To provide a robust platform for studying adversarial attacks and defenses in neural network-based tracking, we also generate 10 videos contain adversarial samples as part of the testing subset using attack toolkit~\cite{jia2020robust}. Therefore, these videos contain additional challenging factor, i.e., AS (influence of Adversarial Samples). A more detailed distribution of each challenge is show in Fig.~\ref{datasetInfo}(b).

\subsection{Statistical Analysis}
The proposed TNLLT dataset comprises 203 unique English words, totaling 1,687 lexical entries, and is specifically designed to focus on the expression of attributes, as shown in Fig.~\ref{datasetInfo}(a). These attributes include spatial locations of objects, relative positions, and other relevant information, as illustrated in the Fig.~\ref{datasetInfo}(b). For the distribution of length of all videos, we can see from Fig.~\ref{datasetInfo}(b) that the TNLLT contains [42,53,74,17,14] videos for the category of less than 1200, 1201-2000, 2001-4000, 4001-6000 and more than 6000. More details, the number of these five segments for train and evaluation set are [30,37,56,14,13] and [12,16,18,3,1] respectively. We can find that the dataset consists of a total of 545,722 frames, with each sequence containing more than 1,000 frames. The longest sequence comprises 14,976 frames, and the average sequence length is 2,728.61 frames. Consequently, this dataset provides a robust foundation for long-term tracking, enabling researchers to thoroughly investigate the challenges associated with extended temporal sequences and develop more resilient tracking algorithms. In this dataset, we have defined 15 challenging factors, among which the four most prevalent challenges are abrupt motion of the camera, scale variation, background clutter and extreme aspect ratio. Videos with these challenging factors will serve as an ideal platform for evaluating the robustness and adaptability of current trackers in the context of long-term tracking. The distribution of the center points of the annotated bounding
boxes is visualized in Fig.~\ref{datasetInfo}(c).

\subsection{Benchmarked Baselines} 
\begin{figure*}[!htp]
\centering
\includegraphics[width=\linewidth]{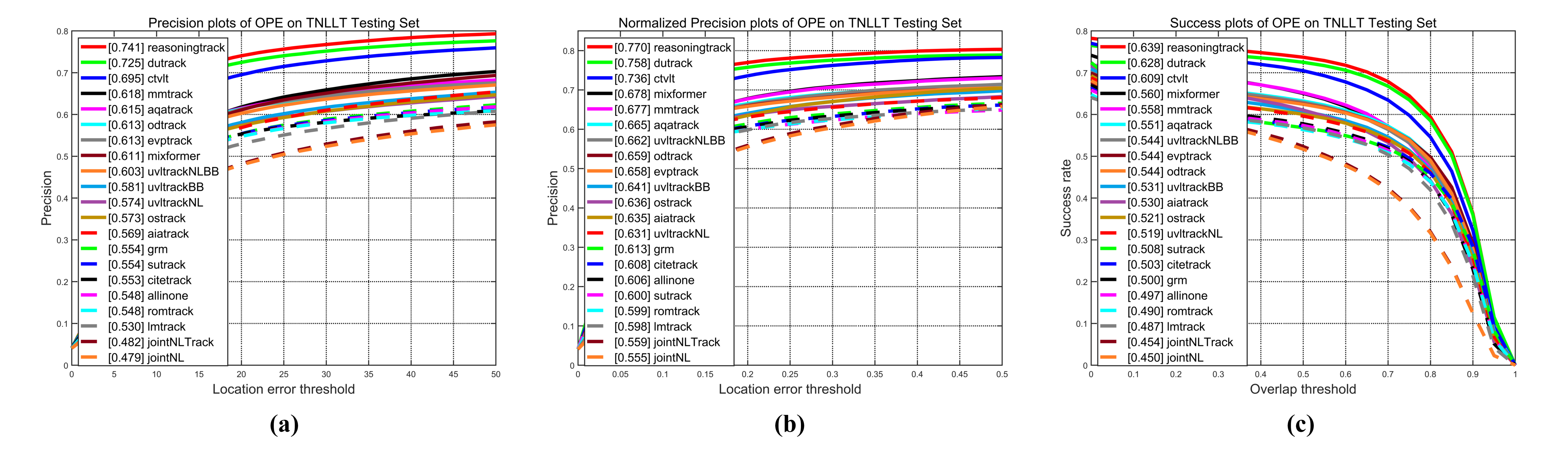}
\caption{Tracking results(PR, NPR, SR) on our newly proposed TNLLT long-term visual language tracking dataset.}
\label{SRPRNPR}
\end{figure*}

\begin{figure*}[!htp]
\centering
\includegraphics[width=\linewidth]{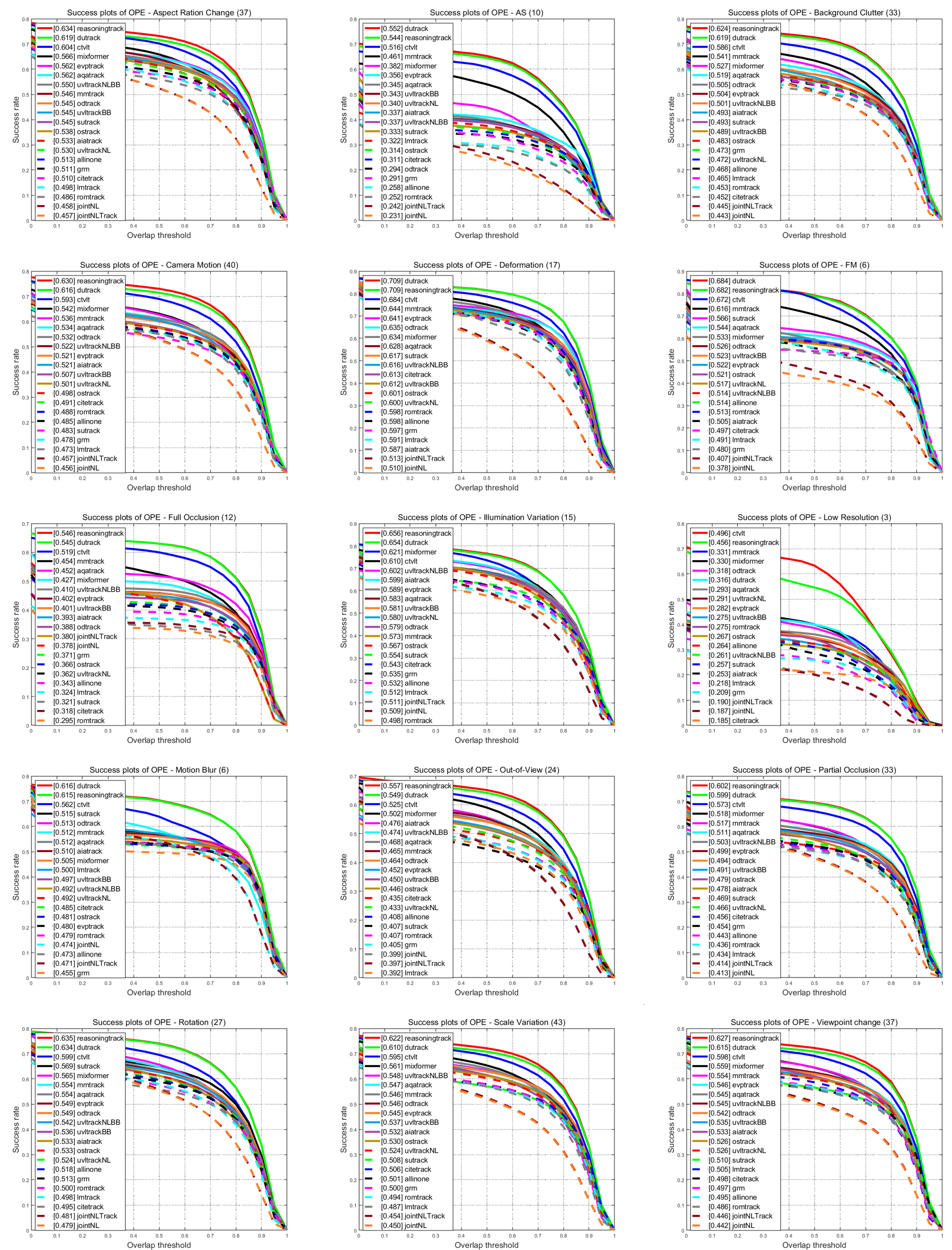}
\caption{Tracking results (SR) under 15 challenging factors.}
\label{challengingfactors}
\end{figure*}

Table~\ref{benchmark} presents a comprehensive comparison of the performance between our proposed ReasoningTrack method and other state-of-the-art tracking approaches. In TNLLT, we choose three types of metrics, Precision, Normalized Precision and Success to evaluate the effectiveness of different tracking algorithms. The quantitative results demonstrate that ReasoningTrack achieves competitive performance compared to mainstream tracking methods across multiple evaluation metrics. As show in Table~\ref{benchmark} and Fig.\ref{SRPRNPR}, our method achieve the best performance across all three metrics, with a PR of 74.1\%, NPR of 77.0\%, and SR of 63.9\%. These results outperform the second-ranked method by 1.6 \%, 1.2\%, and 1.1\%, respectively. Furthermore, as shown in Fig.~\ref{challengingfactors}, current SOTA trackers achieve a relatively low score when meeting attributes like AS, FO, LR and OV. This indicates that these challenging attributes (AS, FO, LR, and OV) still pose significant difficulties for state-of-the-art trackers, leaving considerable room for improvement.

\section{Experiments}  \label{sec::experiments}

\begin{table*}
\center
\small   
\caption{Overall Tracking Performance on TNLLT Dataset. } 
\label{benchmark}
% \resizebox{\textwidth}{!}{ 
\begin{tabular}{l|l|l|l|lll}
\hline \toprule [0.5 pt]
\textbf{No.} & \textbf{Trackers} & \textbf{Source} & \textbf{Type} & \textbf{PR}  &\textbf{NPR}   &\textbf{SR}  \\
\hline
01    &  \textbf{OSTrack~\cite{ye2022ostrack}} & ECCV 2022 & BB & 57.3 & 63.6 & 52.1  \\
02    &  \textbf{MixFormer~\cite{cui2022mixformer}} & CVPR 2022 & BB & 61.1 & 67.8 & 56.0 \\
03    &  \textbf{AiATrack~\cite{gao2022aiatrack}} & ECCV 2022 & BB & 56.9 & 63.5 & 53.0 \\
04    &  \textbf{CiteTrack~\cite{citetracker}} & ICCV 2023 & BB & 55.3 & 60.8 & 50.3 \\
05    &  \textbf{ROMTrack~\cite{cai2023robust}} & ICCV 2023 & BB & 54.8 & 59.9 & 49.0 \\
06    &  \textbf{GRM~\cite{gao2023generalized}} & CVPR 2023 & BB & 55.4 & 61.3 & 50.0 \\
07    &  \textbf{ODTrack~\cite{zheng2024odtrack}} & AAAI 2024 & BB & 61.3 & 65.9 & 54.4   \\
08    &  \textbf{EVPTrack~\cite{shi2024evptrack}} & AAAI 2024 & BB & 61.3 & 65.8 & 54.4 \\
09    &  \textbf{UVLTrack~\cite{ma2024unifying}} & AAAI 2024 & BB & 58.1 & 64.1 & 53.1 \\
10    &  \textbf{AQATrack~\cite{xie2024autoregressive}} & CVPR 2024 & BB & 61.5 & 66.5 & 55.1 \\
11    &  \textbf{LMTrack~\cite{xu2025less}} & AAAI 2025 & BB &53.0 & 59.8 & 48.7 \\

\hline
12    &  \textbf{JointNLT~\cite{zhou2023joint}} & CVPR 2023 & NL & 47.9 & 55.5 & 45.0 \\
13    &  \textbf{JointNLT~\cite{zhou2023joint}} & CVPR 2023 & BL & 48.2 & 55.9 & 45.4 \\
14    &  \textbf{All-in-one~\cite{zhang2023all}} & ACM MM 2023 & BL & 54.8 & 60.6 & 49.7  \\
15    &  \textbf{MMTrack~\cite{zheng2023toward}} & TCSVT 2023 & BL & 61.8 & 67.7 & 55.8 \\
16    &  \textbf{UVLTrack~\cite{ma2024unifying}} & AAAI 2024 & NL & 57.4 & 63.1 & 51.9 \\
17    &  \textbf{UVLTrack~\cite{ma2024unifying}} & AAAI 2024 & BL & 60.3 & 66.2 & 54.4 \\
18    &  \textbf{CTVLT~\cite{feng2025enhancing}} & ICASSP 2025 & BL & 69.5 & 73.6 & 60.9  \\
19    &  \textbf{SUTrack~\cite{sutrack}} & AAAI 2025 & BL & 55.4 & 60.0 & 50.8   \\
20    &  \textbf{DUTrack~\cite{Li2025dutrack}} & CVPR 2025 & BL & 72.5 & 75.8 & 62.8 \\
\hline
21    &  ReasoningTrack (Ours) & - & BL & 74.1 & 77.0 & 63.9  \\

\hline \toprule [0.5 pt]
\end{tabular}
% }
\end{table*}

% \begin{figure}[htbp]
% \centering
% \includegraphics[width=\linewidth]{figures/Precision plots of OPE on TNLLT Testing Set.png} % 自动适应栏宽
% \caption{Visualization of tracking results of our proposed TNLLT dataset.}
% \label{PRSRNPRfig}
% \end{figure}

% \begin{figure}[htbp]
% \centering
% \includegraphics[width=\linewidth]{figures/Precision plots of OPE on LaSOT Testing Set.png} % 自动适应栏宽
% \caption{Visualization of tracking results of Lasot dataset.}
% \label{PRSRNPRfig}
% \end{figure}

\subsection{Implementation Details}
\textbf{Models.}
In this study, we select Qwen2.5 VL 3B as the core base model, as its excellent multimodal fusion ability and cross-modal inference characteristics. In the tracking network module, we adopt DUTrack as our backbone due to its effective fusion of linguistic and visual cues. DUTrack leverages BERT's tokenizer for language processing and employs HiViT's hierarchical patch embedding to encode images into tokens, enabling unified multimodal representation learning. We replaced the text updating module in DUTrack with Qwen2.5 VL, which significantly improved the tracking performance of the model.

\textbf{Training and Inference.}
% 20 baseline visual trackers are re-trained and evaluated on this dataset
In TNLLT benchmark, 20 most recent open-source methods have been retrained on the TNLLT training set, with all training parameters remaining consistent with the original methods.

In the fine-tuning phase of the large model, we perform a two-stage fine-tuning. In the first stage, we sample 1k image training samples from the mainstream visual language tracking dataset, and then use the Qwen-QVQ-Max model to generate chain-of-thought inference data, and each sample was composed of $(x_{S},x{_T},lan,Y)$, where $x_{S}$ and $x_{T}$ represent search frame and template frame respectively, $lan$ denotes the initial language description of the sequence and $Y$ denotes the reasoning responses generated by large language models. Then we use llama factory\cite{zheng2024llamafactory} to perform supervised fine-tuning of the model on the dataset $D_{SFTreason}$ that we collected. Finally, we obtain the model $M_{SFT}$ after supervised fine-tuning. In the second stage, we employ reinforcement learning strategies to further enhance the accuracy of the model's generated language. Specifically, we first sample 1100, 588, 1072, 1118 and 1050 RL training samples from datasets GOT10k, OTB99, TNL2k, LaSOT and TNLLT respectively, with each training sample taking the form of $(box_1,box_2,x_s,x_T,lan)$, where $box_1$ and $box_2$ represent the ground truth of the template frame and the search frame. Then, we fine-tune $M_{SFT}$ using the GRPO algorithm with the reward function we proposed, and the entire training process lasts for a total of 10 epochs. After reinforcement learning, we obtained the final reasoning model $M_{RL}$.

\begin{figure*}
\centering
\includegraphics[width=\linewidth]{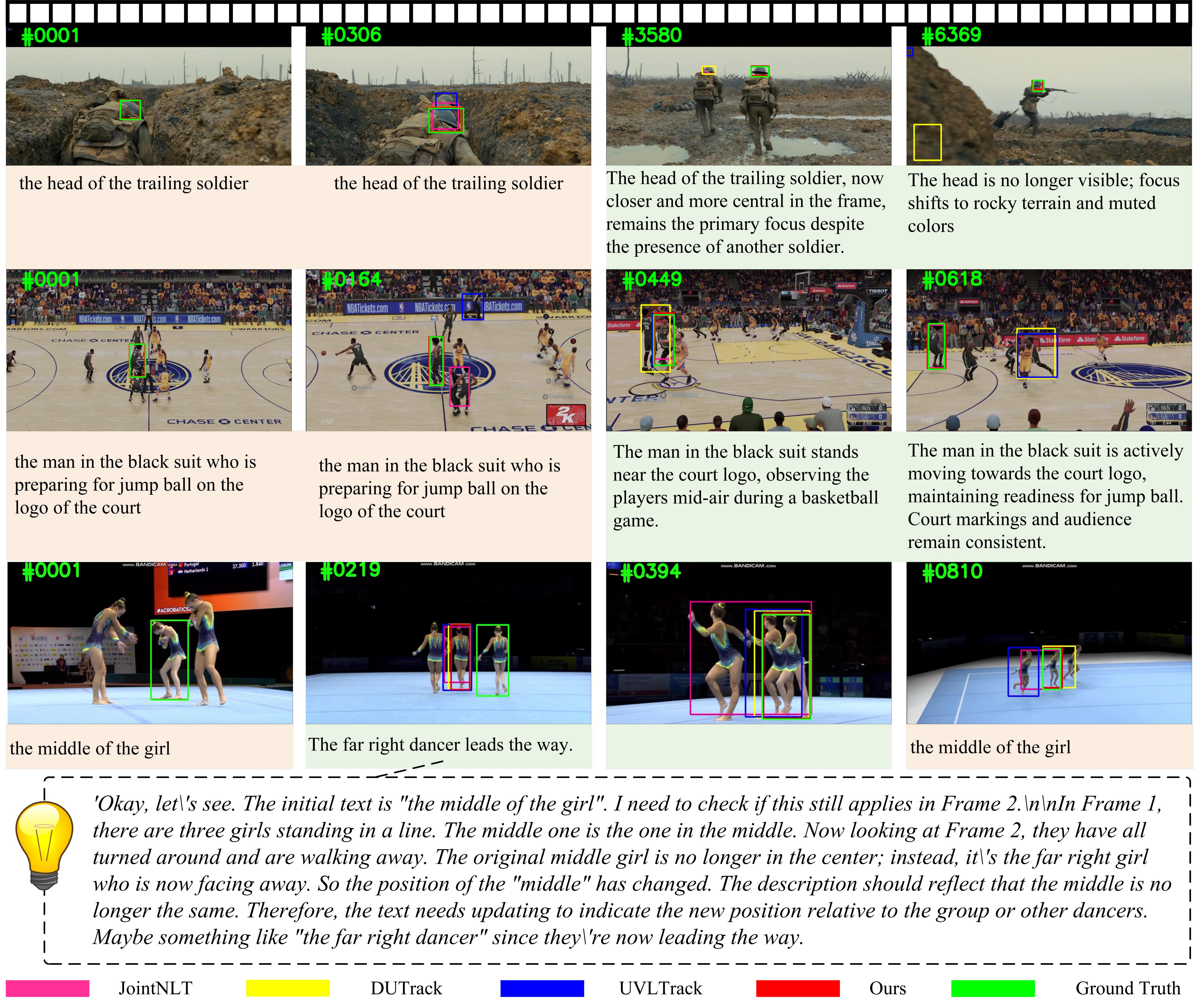} 
\caption{Qualitative comparison results of our tracker VS three VL trackers (i.e., UVLTrack, DUTrack, JointNLT) on three challenging sequences from the TNLLT Benchmark. Each row in the figure represents a video sequence, containing the following components: keyframes sampled along the temporal axis, the original textual description, the refined textual description, the chain-of-thought reasoning process, and the video object tracking results.}
\label{visiualization}
\end{figure*}

During the testing phase, we replace the text update module of our baseline DUTrack while maintaining its original multi-modal interaction capabilities and template frame updating abilities. Specifically, first,  we deploy our fine-tuned model $M_{RL}$ using vllm\cite{kwon2023efficient}, then we set an update interval $u$, and the text update module is invoked once every $u$ frames. In order to reduce the overhead of invoking LLM, we follow the approach of DUTrack by using a function $f$ to preliminarily determine whether a text update is necessary. Then, if the model outputs a signal indicating that an update is needed within the $<d>$ tag, we will extract the results of the model's updated text from the $<answer>$ tag and input them into the backbone network to complete the subsequent tracking process. Algorithm~\ref{alg:llm_tracking} describes in detail the overall process of the testing phase. It is worth noting that since the test set of the GOT10k does not provide any language information, we utilize the language annotations provided in Guo's work~\cite{guo2024divert} to complete the text updates. All training and testing were conducted on A800 GPUs.

\subsection{Dataset and Evaluation Metric}
\noindent $\bullet$ \textbf{OTB99}~\cite{li2017tracking} is an extension of OTB100. To explore the potential of language-guided tracking, textual annotations were added to OTB100 dataset. OTB99 consists of 99 video sequences, divided into 51 for training and 48 for testing. As shown in Table~\ref{OTB99_results}, our method achieves the best performance in two metrics, with an AUC of 71.11\% and PR of 95.58\%. These results indicate that ReasoningTrack can significantly enhance the tracking performance in short-term tracking.

\noindent $\bullet$ \textbf{GOT-10k}~\cite{huang2019got} includes 10,000 videos covering 563 categories of objects, encompassing a rich array of moving targets in real-world scenarios. In order to explore the impact of text on tracking performance, we utilized the text annotations from Guo's work~\cite{guo2024divert}. The experimental results presented in Table~\ref{GOT10k_results} indicate that after introducing the strategy of COT text update mechanism, our method achieves competitive results ($AO$:77.8\%, $SR_{0.5}$:88.5, $SR_{0.75}$:77.0) on this dataset, significantly enhancing the performance of the baseline tracker.

\noindent $\bullet$ \textbf{TNL2K}~\cite{wang2021tnl2k} dataset is designed for natural language-based tracking, which consists of 2000 video sequences collected from YouTube, intelligent surveillance cameras, and mobile phones. The dataset is divided into 1300 training videos and 700 test videos. As shown in Table~\ref{TNL2k_results}, our method achieves competitive performance, improving by 3\% in PR over the second-ranked method. These results demonstrate its effectiveness in natural language-based tracking.

\noindent $\bullet$ \textbf{TNLLT} dataset is an extension of the TNL2K dataset, which contains 200 long-term videos, and the dataset consists of a total of 545,722 frames, with each sequence containing more than 1,000 frames. The longest sequence comprises 14,976 frames and the average sequence length is 2,728.61 frames. The dataset is evaluated using three metrics: precision (P), normalize precision(NP) and success rate(SR) metrics. As shown in Table~\ref{benchmark}, our method has achieved the best results among the various approaches in recent years, with an PR of 74.1\%, NPR of 77.0\%, SR of 63.9\%.

% $\bullet$ \textbf{Results on GOT-10k dataset.} 
\begin{table}
\center
\small   
\caption{Overall tracking performance on GOT-10k dataset. } 
\label{GOT10k_results}
% \resizebox{\columnwidth}{!}{ 
\begin{tabular}{l|cccccc}
\hline 
\textbf{Trackers}    &\textbf{AO} & \textbf{SR$_{0.5}$}  &\textbf{SR$_{0.75}$}    \\
\hline
\textbf{ SiamFC}\cite{bertinetto2016fully}               &34.8  &35.3       &9.8                \\ 
\textbf{ MDNet} \cite{zhang2017mdnet}              &29.9  &30.3       &9.9                \\ 
\textbf{ Ocean}\cite{zhang2020ocean}               &61.1  &-       &-                \\ 
\textbf{ SiamPRN++}\cite{li2019siamrpn++}               &51.7  &-       &-                \\ 
\textbf{ TrDimp} \cite{wang2021transformer}              &68.8  &80.5       &59.7                \\ 
\textbf{ TransT} \cite{chen2021transformer}              &72.3  &82.4       &68.2                \\ 
\textbf{ SimaTrack} \cite{luo2021exploring}              &68.6  &-       &-                \\ 
\textbf{ VideoTrack}\cite{xie2023videotrack}               &72.9  &-       &-                \\ 
\textbf{ SeqTrack}\cite{chen2023seqtrack}               &74.7  &84.7       &71.8                \\ 
\textbf{ ODTrack} \cite{zheng2024odtrack}              &77.0  &-       &-                \\ 
\textbf{ MixFormer}  \cite{cui2022mixformer}             &75.6  &85.7       &72.8                \\
\textbf{ ROMTrack-384} \cite{cai2023robust}              &74.2  &84.3       &72.4                \\ 
\textbf{ DUTrack}\cite{Li2025dutrack}              &76.7  &-       &-                \\ 
\textbf{ VLT-384} \cite{guo2024divert}              &76.3  &87.1       &72.3                \\ 
\textbf{ AQATrack} \cite{xie2024autoregressive}              &76.0  &85.2       &74.9                \\ 
\textbf{ TemTrack-384} \cite{xie2025robust}              &76.1  &84.9       &74.4                \\ 
\textbf{ MambaLCT-384} \cite{li2025mambalct}              &76.2  &86.7       &74.43               \\ 
\hline
\textbf{ Ours }         & 77.8  & 88.5   & 77.0  \\
\hline
\end{tabular}
% }
\end{table}

% $\bullet$ \textbf{Results on OTB99 dataset.} 
\begin{table}
\center
\small   
\caption{Overall tracking performance on OTB99 dataset. } 
\label{OTB99_results}
% \resizebox{\columnwidth}{!}{ 
\begin{tabular}{l|c|ccccc}
\hline 
\textbf{Trackers} & \textbf{Source}   & \textbf{PR}   &\textbf{AUC} \\
\hline
\textbf{ LSTMTrack}\cite{feng2020real}    &WACV 2020           &79.0       &61.0           \\ 
\textbf{ SNLT}\cite{feng2021siamese}    &CVPR 2021           &84.8       &66.6            \\ 
\textbf{ GTI}\cite{yang2020grounding}    &TCSVT 2021           &73.2       &57.1          \\ 
\textbf{ TNL2k-II}\cite{wang2021tnl2k}    &CVPR 2021           &88.0       &68.0           \\ 
\textbf{ TransVLT}\cite{zhao2023transformer}    &PRL 2023           &91.2       &69.9          \\ 
\textbf{ JointNLT}\cite{zhou2023joint}    &CVPR 2023           &85.6       &65.3           \\ 
\textbf{ MMTrack-384}\cite{zheng2023toward}    &TCSVT 2023          &91.8       &70.5            \\ 
\textbf{ ATTrack}\cite{ge2024consistencies}    &MM 2024          &90.3       &69.3        \\ 
\textbf{ OSDT}\cite{zhang2024one}    &TCSVT 2024          &86.7       &66.2            \\ 
\textbf{ UVLTrack-B}\cite{ma2024unifying}    &AAAI 2024          &89.9       &69.3           \\ 
\textbf{ QueryNLT}\cite{shao2024context}    &CVPR 2024          &88.2       &66.7            \\ 
\textbf{ DUTrack}\cite{Li2025dutrack}    &CVPR 2025          &93.9       &70.9           \\ 
\textbf{ ATSTrack}\cite{zhen2025atstrack}    &arxiv 2025          &94.4       &71.0           \\ 
\hline
\textbf{ Ours }   & -      & 95.58  & 71.11    \\
\hline
\end{tabular}
% }
\end{table}

\begin{table}
\center
\small   
\caption{Overall tracking performance on TNL2K dataset. } 
\label{TNL2k_results}
% \resizebox{\columnwidth}{!}{ 
\begin{tabular}{l|c|cccc}
\hline 
\textbf{Trackers} & \textbf{Source}   & \textbf{AUC}     &\textbf{PR} \\
\hline
\textbf{OSTrack}\cite{ye2022ostrack}    &ECCV 2022           &55.9          &-        \\  
\textbf{SeqTrack}\cite{chen2023seqtrack}    &CVPR 2023          &54.9          &-      \\ 
\textbf{AQATrack}\cite{xie2024autoregressive}    &CVPR 2024           &57.8            &59.4       \\ 
\textbf{ODTrack-384}\cite{zheng2024odtrack}    &AAAI 2024          &60.9           &-         \\ 
\textbf{JointNLT}\cite{zhou2023joint}    &CVPR 2023          &56.9           &58.1        \\ 
\textbf{MMTrack-384}\cite{zheng2023toward}    &TCSVT 2023           &58.6             &59.4         \\ 
\textbf{ATTrack}\cite{ge2024consistencies}    &MM 2024          &56.9            &64.7         \\ 
\textbf{UVLTrack}\cite{ma2024unifying}    &AAAI 2024          &62.7             &65.4       \\ 
\textbf{OSDT}\cite{zhang2024one}    &TCSVT 2024           &59.3             &61.5         \\ 
\textbf{QueryNLT}\cite{shao2024context}    &CVPR 2024          &57.8             &58.7         \\ 
\textbf{DUTrack}\cite{Li2025dutrack}    &CVPR 2025           &64.9          &70.6     \\ 
\textbf{ATSTrack}\cite{zhen2025atstrack}    &arxiv 2025           &66.2           &71.3   \\
\textbf{TemTrack}\cite{xie2025robust}    &AAAI 2025           &58.8           &-   \\
\hline
\textbf{ Ours }   & -      & 65.3     &  74.3 \\
\hline
\end{tabular}
% }
\end{table}
% $\bullet$ \textbf{Results on LaSOT-{ext} dataset.} 

% \begin{table}
% \center
% \small   
% \caption{Overall tracking performance on LaSOT-{ext} dataset. } 
% \label{OTB100_results}
% % \resizebox{\columnwidth}{!}{ 
% \begin{tabular}{l|c|ccccc}
% \hline 
% \textbf{Trackers} & \textbf{Source}   & \textbf{AUC}  &\textbf{NPR}   &\textbf{PR} \\
% \hline
% \textbf{-}    &- &-       &-      &-          \\  
% \textbf{-}    &- &-       &-      &-         \\  
% \hline
% \textbf{ Ours }   & -      & -  & -   & -  \\
% \hline
% \end{tabular}
% % }
% \end{table}

% $\bullet$ \textbf{Results on TNL2K dataset.} 

% \begin{table}
% \center
% \small   
% \caption{Overall tracking performance on TNL2K dataset. } 
% \label{OTB100_results}
% % \resizebox{\columnwidth}{!}{ 
% \begin{tabular}{l|c|ccccc}
% \hline 
% \textbf{Trackers} & \textbf{Source}   & \textbf{AUC}  &\textbf{PR}   &\textbf{NPR} \\
% \hline
% \textbf{-}    &-         &-      &-      &-          \\  
% \hline
% \textbf{ - }   & -      & -  &-   & -  \\
% \hline
% \end{tabular}
% % }
% \end{table}

% $\bullet$ \textbf{Results on TNLLT dataset.} 

\subsection{Ablation Study} 
\noindent $\bullet$ \textbf{Analysis on Two-stage Finetuning Strategy.} To better demonstrate the effectiveness of our module, we conduct ablation experiments on the TNLLT and OTB99 benchmarks. As shown in Table~\ref{absftrl}, we investigate the impact of our two-stage fine-tuning strategy on the final tracking accuracy. It can be observed that supervised fine-tuning boosts the tracker’s accuracy, and subsequent reinforcement learning yields a further precision gain, confirming that our two-stage pipeline effectively enhances the model’s ability to generate accurate updated text.

\noindent $\bullet$ \textbf{Analysis on Reward Function.} In Table ~\ref{abreward}, we further explore the impact of the two main reward signals in our reinforcement learning process on tracking accuracy.  After two rounds of reinforcement learning, the experimental results indicate that both IoU Reward (IR) and Judge Reward (JR) improve the accuracy of text updates to some extent, thereby enhancing the precision of the tracking algorithm.

\begin{figure}[!htp]
\centering
\includegraphics[width=\linewidth]{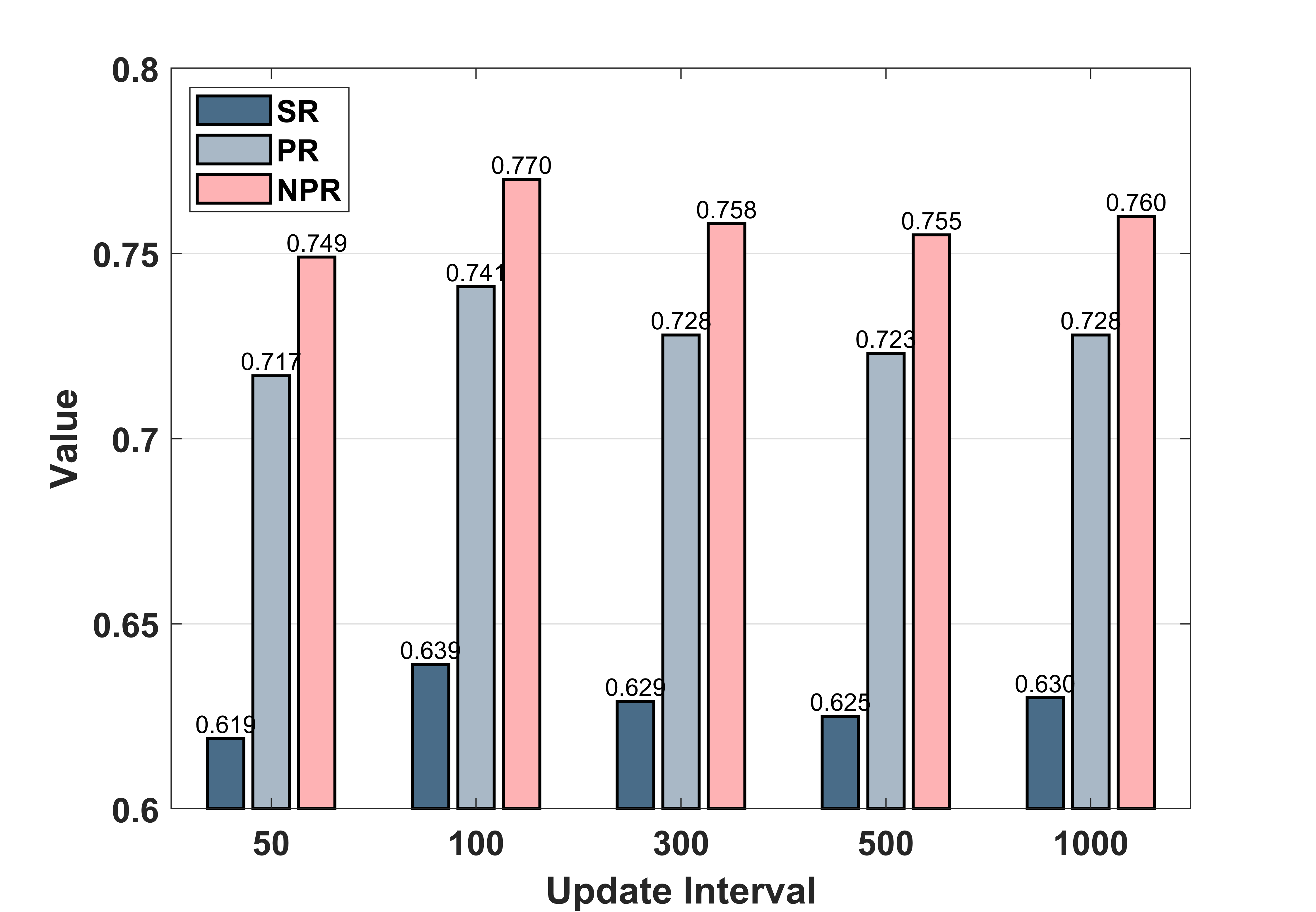}
\caption{Adjustment of update interval in Testing Phase.}
\label{update_interval}
\end{figure}

% \begin{table}
% \center
% \small     
% \caption{updating text frequency in TNLLT dataset.} 
% \label{eventvot_ablation}
% \resizebox{\columnwidth}{!}{ 
% \begin{tabular}{l|llll} 
% \hline 
% \textbf{\# update interval}   &\textbf{PR}   & \textbf{NPR} &\textbf{SR} \\
% \hline
% \text{50}         & 71.7  & 74.9  & 61.9 \\
% \text{100}         & 74.1  & 77.0  & 63.9 \\
% \text{300}         & 72.8  & 75.8  & 62.9 \\
% \text{500}         & 72.3  & 75.5  & 62.5 \\
% \text{1000}         & 72.8  & 76.0  & 63.0 \\
% \hline
% \end{tabular}
% }
% \end{table}
% **OTB99lan actor130\_interval50 TNLLT\_actor90\_interval300** dyanmic1 means use updated lan to update, dynamic2 menas use static lan to update
\begin{table}
\center
\small     
\caption{Ablation studies of static and dynamic text infomation on TNLLT.} 
\label{updateway_ablation}
\resizebox{0.9\columnwidth}{!}{ 
\begin{tabular}{l|l|lll} 
\hline 
\multicolumn{1}{c|}{\textbf{Dataset}} & \textbf{\# Settings} & \textbf{PR} & \textbf{NPR} & \textbf{SR} \\
\hline 
% \multicolumn{1}{c|}{\multirow{5}{*}{OTB99lan}} & static         & 95.48  & 87.31  & 71.08  \\
% \multicolumn{1}{c|}{} & dynamic1       & 95.12  & 87.22  & 70.89  \\
% \multicolumn{1}{c|}{} & dynamic2       & 94.65  & 86.85  & 70.84  \\
% \multicolumn{1}{c|}{} & static+dynamic & 94.73  & 87.03  & 70.84  \\
% \multicolumn{1}{c|}{} & dynamic+static & 94.73  & 86.93  & 70.87  \\
% \hline 
\multicolumn{1}{c|}{\multirow{4}{*}{TNLLT}} & static         &  72.6 &  75.8 &   62.8\\
\multicolumn{1}{c|}{} & dynamic1       & 72.8  & 76.1  &  63.0 \\
\multicolumn{1}{c|}{} & dynamic2       & 72.8  & 75.8  &  62.9 \\
% \multicolumn{1}{c|}{} & static+dynamic &  72.5 & 75.9  &  62.9 \\
\multicolumn{1}{c|}{} & dynamic+static & \textbf{73.1}  & \textbf{76.3}  &  \textbf{63.2} \\
\hline 
\end{tabular}
}
\end{table}

\begin{figure*}[!htp]
\centering
\includegraphics[width=\linewidth]{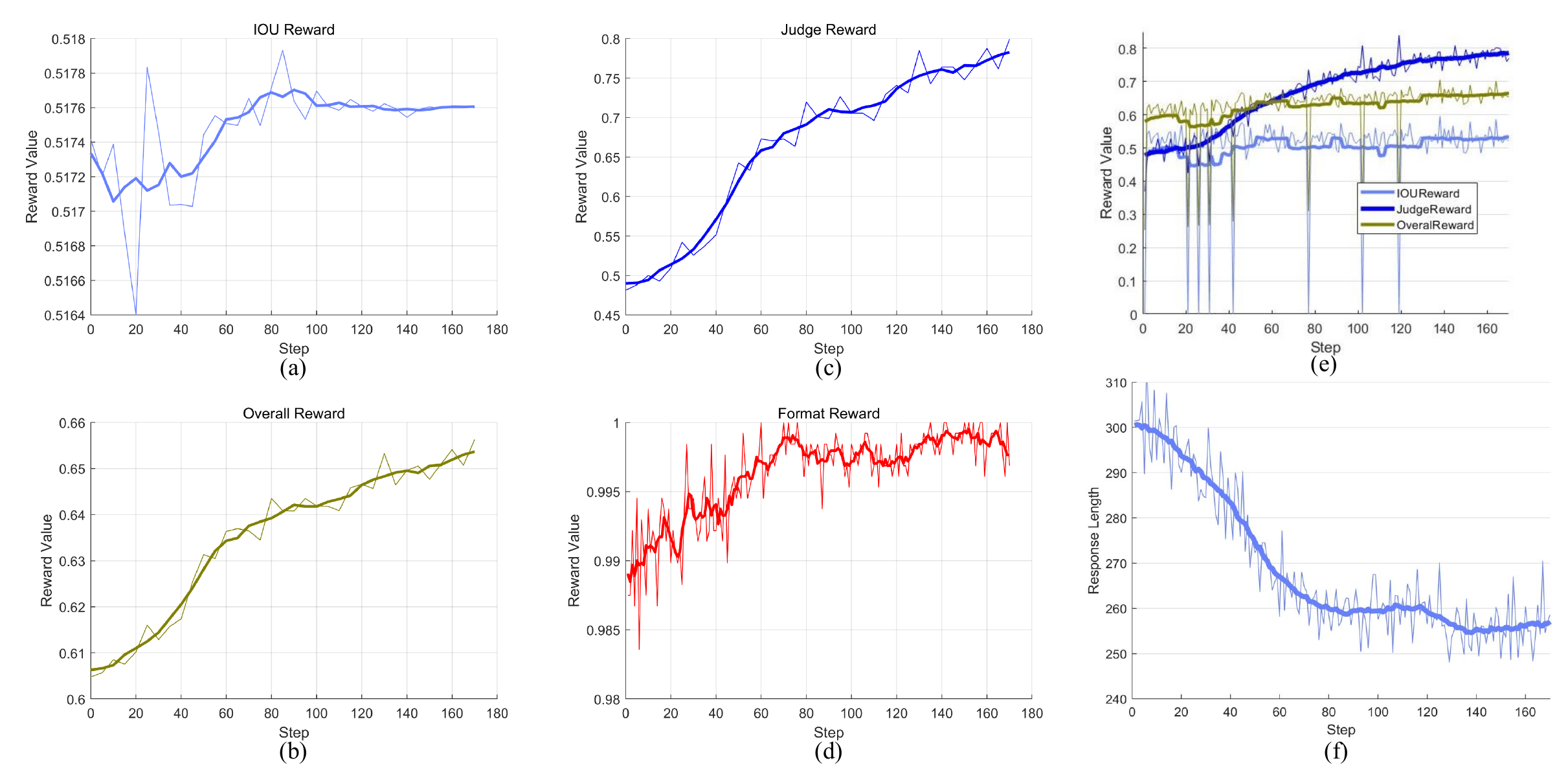}
\caption{Reward values during the validation phase of reinforcement learning (a-d), the trend of the reward signals (e) and the trend of the response length (f) in reinforcement learning training phase. }
\label{valReward}
\end{figure*}

% \begin{table}
% \centering
% \small     
% \caption{Ablation studies of SFT and RL.} 
% \label{eventvot_ablation}
% \resizebox{\columnwidth}{!}{ 
% \begin{tabular}{l|l|lll} 
% \hline 
% \textbf{Dataset} & \textbf{\# Settings} & \textbf{PR} & \textbf{NPR} & \textbf{SR} \\
% \hline
% \multirow{3}{*}{TNLLT} & \text{baseline} & 66.7 & 70.4 & 62.8 \\
% & \text{+sft} & 67.3 & 70.9 & 63.2 \\
% & \text{+sft+rl} & 68.2 & 71.5 & 63.9 \\
% \hline
% \multirow{3}{*}{OTB99lan} & \text{baseline} & 93.9 & - & 70.9 \\
% & \text{+sft} & 95.17 & - & 70.99 \\
% & \text{+sft+rl} & 95.58 & - & 71.11 \\
% \hline
% \end{tabular}
% }
% \end{table}

%%%%%%%%%%%%%%%%%%%%%%%%%%%%%%%%%%%%%%%%%%%%%%%%%%%%%%%%%%%%%%%%%%%%%%
\begin{table}
\centering
\small
\caption{Ablation study for SFT and RL } 
\begin{tabular}{c|cc|ccc|cc}
\hline 
\multirow{2}{*}{\textbf{\#No.}} & \multirow{2}{*}{\textbf{SFT}} & \multirow{2}{*}{\textbf{RL}}  & \multicolumn{3}{c|}{\textbf{TNLLT}} & \multicolumn{2}{c}{\textbf{OTB99}} \\
\cline{4-8}
& & & \textbf{PR} & \textbf{NPR} & \textbf{SR} & \textbf{PR}  & \textbf{SR} \\
\cline{1-8}
1 & &  & 72.5 & 75.8 & 62.8 & 93.90  & 70.90  \\
% \cline{1-10}
2 & $\checkmark$ &  & 73.1 & 76.4 & 63.2 & 95.17  & 70.99  \\
% \cline{1-10}
3 & $\checkmark$ & $\checkmark$ &\textbf{74.1} & \textbf{77.0}  & \textbf{63.9} & \textbf{95.58}  & \textbf{71.11}  \\
\hline 
\end{tabular} 
\label{absftrl}
\end{table}
%%%%%%%%%%%%%%%%%%%%%%%%%%%%%%%%%%%%%%%%%%%%%%%%%%%%%%%%%%%%%%%%%%%%%%

%%%%%%%%%%%%%%%%%%%%%%%%%%%%%%%%%%%%%%%%%%%%%%%%%%%%%%%%%%%%%%%%%%%%%%

\begin{table}
\centering
\small
\caption{Ablation study for RL reward.} 
\begin{tabular}{c|cc|ccc|cc}
\hline 
\multirow{2}{*}{\textbf{\#No.}}  & \multirow{2}{*}{\textbf{IR}} & \multirow{2}{*}{\textbf{JR}} & \multicolumn{3}{c|}{\textbf{TNLLT}} & \multicolumn{2}{c}{\textbf{OTB99}} \\
\cline{4-8}
& & & \textbf{PR} & \textbf{NPR} & \textbf{SR} & \textbf{PR}  & \textbf{SR} \\
\cline{1-8}
1  & $\checkmark$ & \null & 73.2 & 76.4 & 63.3 & 94.71 & 70.81 \\ 
% \cline{1-10}
2  & \null & $\checkmark$ & 73.0 & 76.1 & 63.1 & 95.11 & \textbf{70.95} \\
% \cline{1-10}
3  & $\checkmark$ & $\checkmark$ & \textbf{73.4} & \textbf{76.6} & \textbf{63.4} & \textbf{95.20} & 70.91 \\
\hline 
\end{tabular} 
\label{abreward}
\end{table}
%%%%%%%%%%%%%%%%%%%%%%%%%%%%%%%%%%%%%%%%%%%%%%%%%%%%%%%%%%%%%%%%%%%%%%

% \subsection{Component Analysis} 
\noindent $\bullet$ \textbf{Component Analysis on Update Interval.} As shown in Fig.~\ref{update_interval}, we investigate the impact of text update frequency on tracking accuracy. Specifically, we set the update frequency to 50, 100, 300, 500, 1000, respectively and find that the tracking accuracy is highest when the text update frequency is 100.

\begin{figure*}[!htp]
\centering
\includegraphics[width=\textwidth]{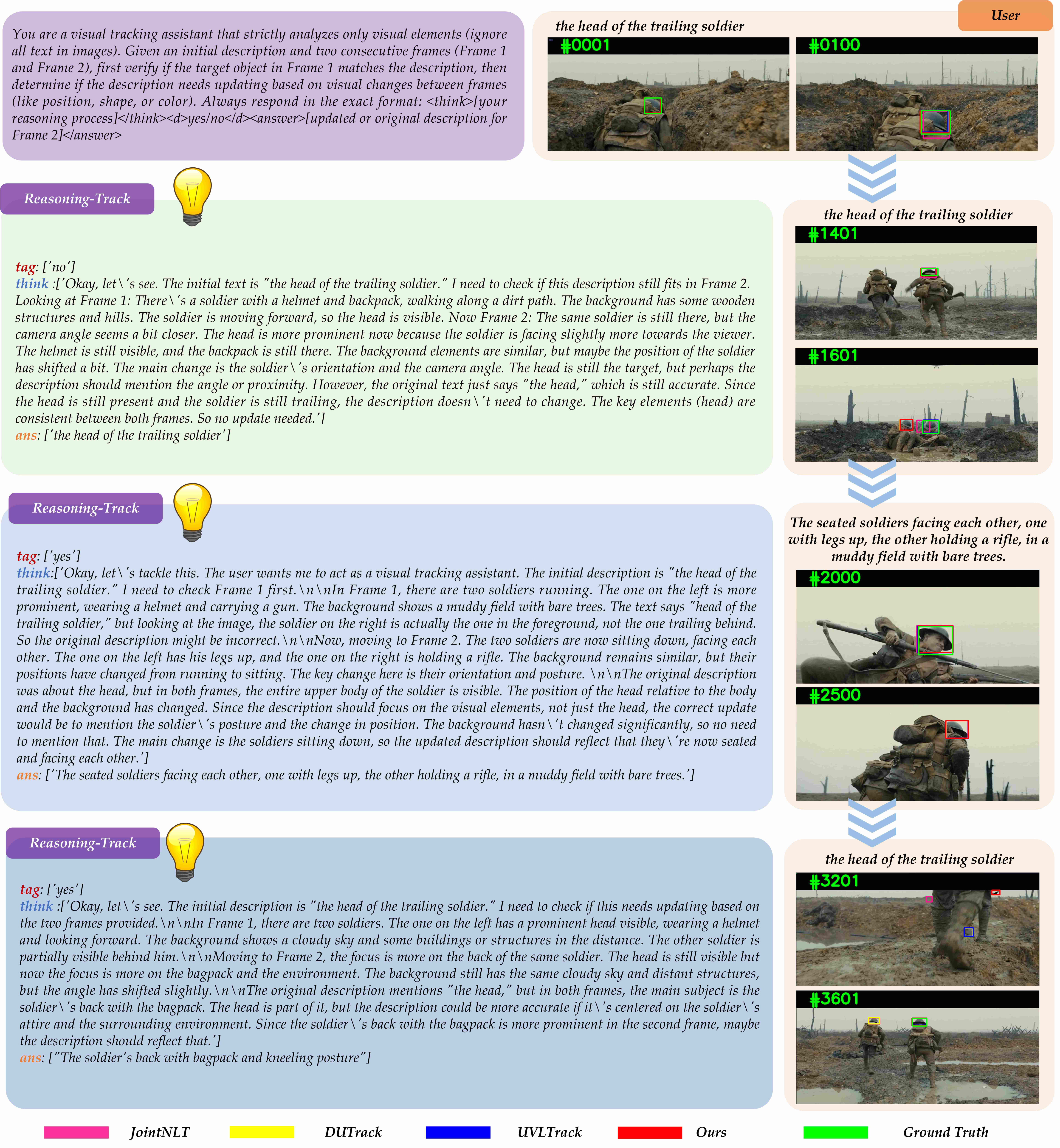}
\caption{More examples1.}
\label{more_examples1}
\end{figure*}

\begin{figure*}[!htp]
\centering
\includegraphics[width=\textwidth]{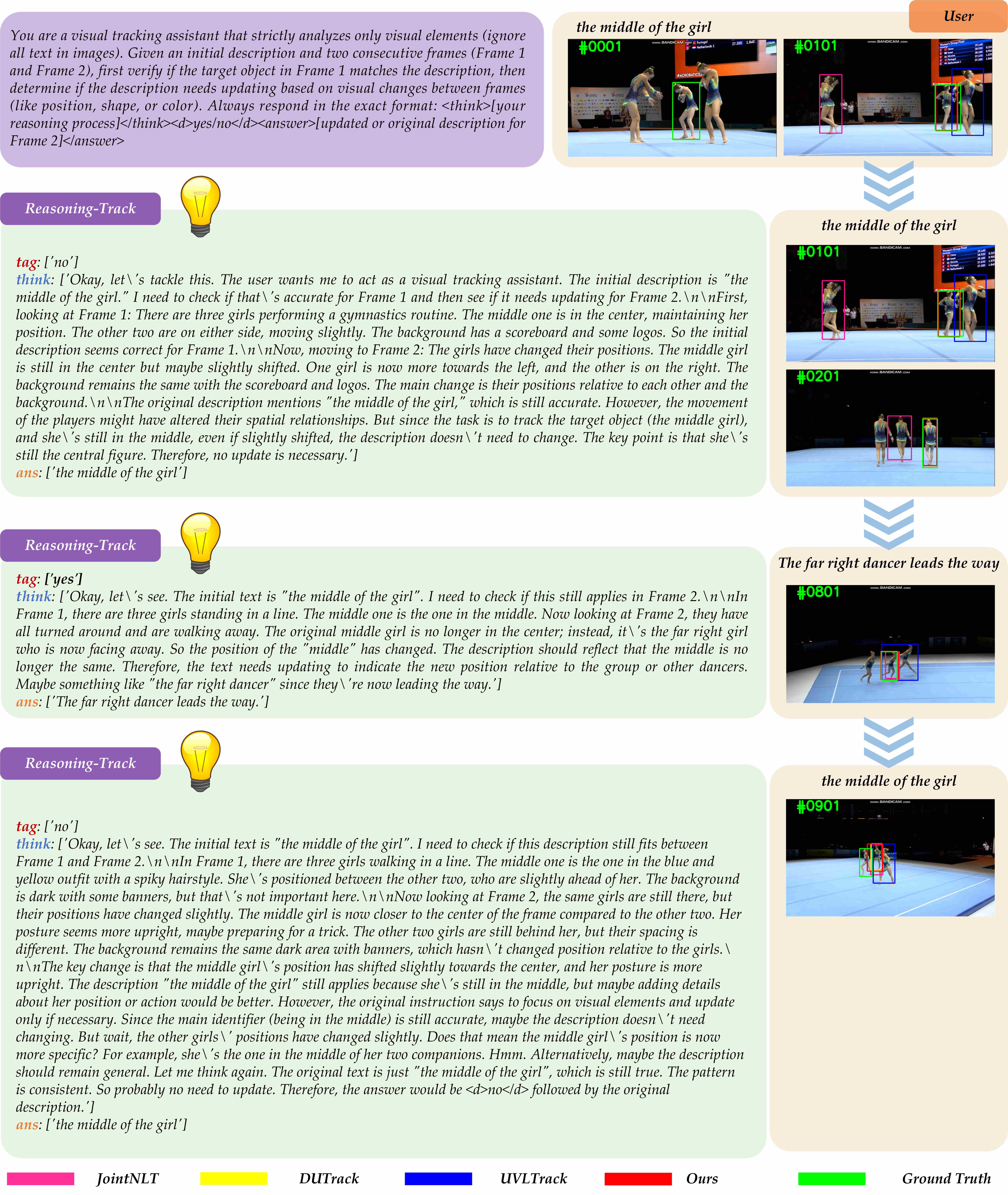}
\caption{More examples2.}
\label{more_examples2}
\end{figure*}

\begin{figure*}[!htp]
\centering
\includegraphics[width=\textwidth]{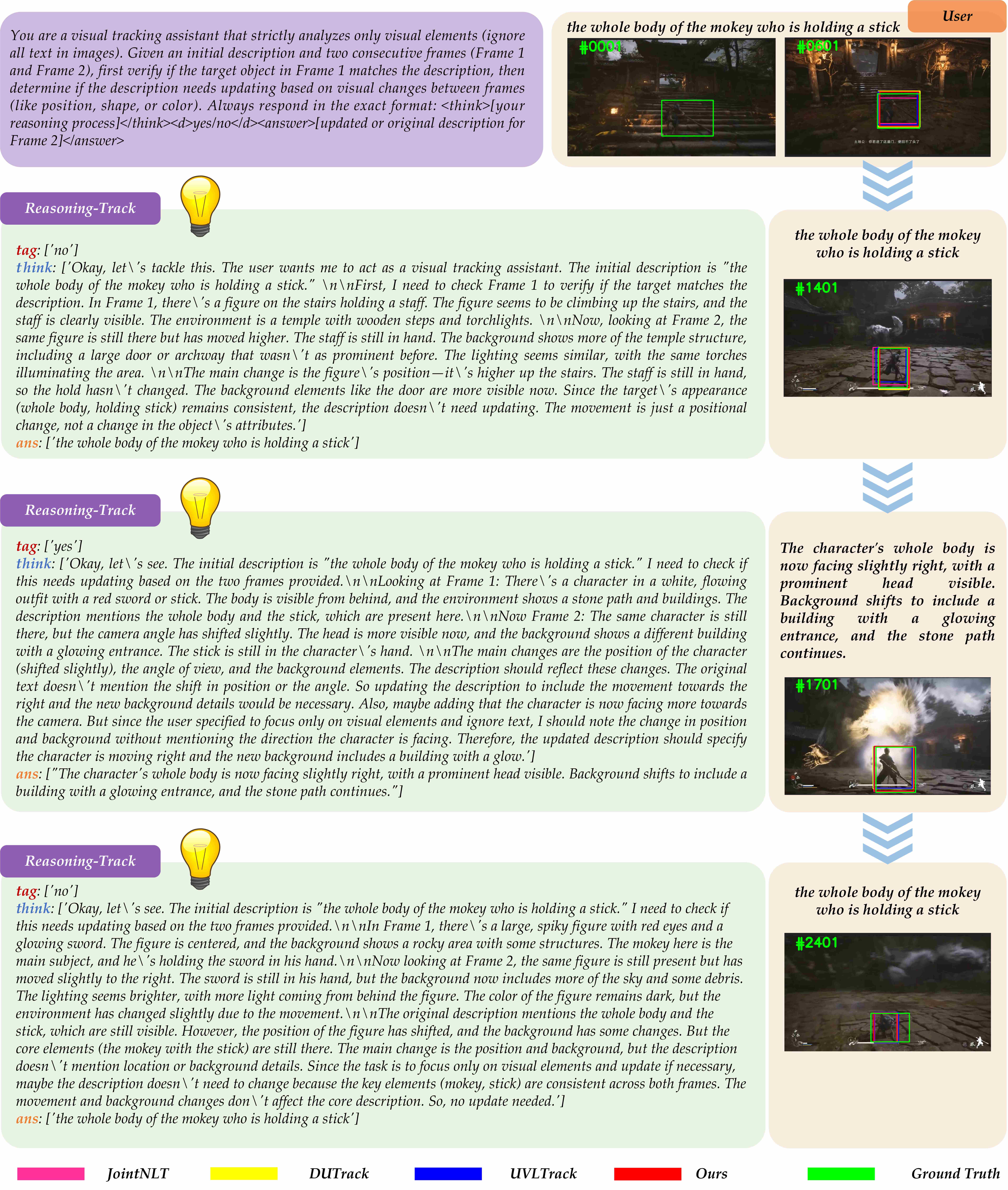}
\caption{More examples3.}
\label{more_examples3}
\end{figure*}

\noindent $\bullet$ \textbf{Component Analysis on Text Update Strategy.} In Table~\ref{updateway_ablation}, we compare the impact of different text update strategies on tracking accuracy. In this part of the experiment, we fixed the interval for text updates at 300. The `static' configuration parameter specifies the use of static text throughout the tracking process, while the `dynamic1' setting refers to the text description input for the large model comes from the initial text provided in the dataset. `dynamic2' denotes that the text input to the large model is derived from the results of the large model's cascading updates. `dynamic+static' refers to the sequential concatenation of dynamic and static text input into the backbone network. We can find that after the introduction of dynamic text updates, there has been a consistent improvement in the model's performance. Cascaded updates are slightly lower in accuracy compared to updates using the initial text description, which may be due to error accumulation. Furthermore, integrating dynamic and static languages as inputs to the language branches of the backbone network can further enhance the model's tracking accuracy, potentially due to the more robust information contained within the static language.

\subsection{Visualization}

As shown in the qualitative comparison results in Fig.~\ref{visiualization}, the initial textual description can easily mislead the subsequent tracking process. In the third video sequence, the initial text description is `the middle of the girl', however, in frame 219, the girl in the middle move to the far right, which no longer aligns with the initial text. After employing our text refinement mechanism, the initial text becomes `The far right dancer leads the way', which aligns with the characteristics of the image. These refined text descriptions can effectively enhance the accuracy of long-term tracking and improve the robustness of tracking algorithms in different scenarios. Compared to existing tracking methods, our approach not only offers dynamic text but also provides the reasoning processes, which offer stronger interpretability for text updates. We also demonstrates the model's text-update trajectories in several typical scenarios, visualizing its reasoning process and tracking results, as detailed in Fig.\ref{more_examples1}, Fig.\ref{more_examples2} and Fig.\ref{more_examples3}. 
% Extensive experiments on multiple datasets indicate that our approach significantly enhances the accuracy of text updates and improves the robustness of tracking algorithms in complex scenarios. 

In order to analyze the training dynamics of our reinforcement learning process, we monitor key indicators that influence the convergence of reinforcement learning, illustrating the evolution of reward signals and model behavior throughout the entire training process.

As shown in Fig.\ref{valReward} (a-d), this figure tracks the changes in different reward signals during RL training, evaluated on the validation set. We conduct training over 10 epochs, consisting of 170 steps in total. This curve indicates that the reward signal that we designed continues to increase throughout the training process, demonstrating the stability and convergence of the training process. As shown in Fig.\ref{valReward} (e), we also monitored the changes in the reward signals during the training process. Furthermore, in Fig.\ref{valReward} (f), we observe that the response length of the model has decreased from an initial 300 to approximately 255, indicating that as reinforcement learning progressed, the model learned to respond in a more concise and efficient way.

\section{Limitation Analysis} 
Although our approach demonstrates strong performance across multiple benchmarks, it has some notable limitations. First, the integration of large pre-trained models incurs significant computational costs, resulting in a tracking speed of only 15.57 fps. This frame rate may be insufficient for real-time applications requiring high-speed processing. Second, our fixed-interval text update strategy cannot dynamically adapt to rapid appearance changes. Furthermore, as shown in the second sequence of Fig.~\ref{visiualization}, our updated text precisely indicates the dancer on the right side, but the tracker, maybe influenced by the historical process, still failed to make an accurate identification.

\section{Conclusion}  \label{sec::Conclusion}
In this paper, we propose ReasoningTrack, a novel natural language reasoning strategy for vision-language tracking based on pre-trained large vision-language models. It employs a two-stage strategy of supervised fine-tuning and reinforcement learning to refine the LLM in order to optimize the method for describing the tracking process in text. In addition, the proposed method is a plug-and-play method that can be seamlessly integrated into existing VLTracking methods to improve tracking accuracy. Furthermore, in order to address the issue of lacking long-term tracking datasets, we propose a large-scale long-term vision-language tracking benchmark TNLLT, which builds a solid foundation for the vision-language visual tracking task. Future research may need to focus on simplifying the large model framework to develop text update methods that are both efficient and accurate and explore more effective methods of integrating textual cues into tracking frameworks.

% \newpage

\small{ 
\bibliographystyle{IEEEtran}
\bibliography{reference}

\begin{thebibliography}{10}
\providecommand{\url}[1]{#1}
\csname url@samestyle\endcsname
\providecommand{\newblock}{\relax}
\providecommand{\bibinfo}[2]{#2}
\providecommand{\BIBentrySTDinterwordspacing}{\spaceskip=0pt\relax}
\providecommand{\BIBentryALTinterwordstretchfactor}{4}
\providecommand{\BIBentryALTinterwordspacing}{\spaceskip=\fontdimen2\font plus
\BIBentryALTinterwordstretchfactor\fontdimen3\font minus
  \fontdimen4\font\relax}
\providecommand{\BIBforeignlanguage}[2]{{%
\expandafter\ifx\csname l@#1\endcsname\relax
\typeout{** WARNING: IEEEtran.bst: No hyphenation pattern has been}%
\typeout{** loaded for the language `#1'. Using the pattern for}%
\typeout{** the default language instead.}%
\else
\language=\csname l@#1\endcsname
\fi
#2}}
\providecommand{\BIBdecl}{\relax}
\BIBdecl

\bibitem{zhou2023joint}
L.~Zhou, Z.~Zhou, K.~Mao, and Z.~He, ``Joint visual grounding and tracking with
  natural language specification,'' in \emph{Proceedings of the IEEE/CVF
  conference on computer vision and pattern recognition}, 2023, pp.
  23\,151--23\,160.

\bibitem{ma2024unifying}
Y.~Ma, Y.~Tang, W.~Yang, T.~Zhang, J.~Zhang, and M.~Kang, ``Unifying visual and
  vision-language tracking via contrastive learning,'' 2024.

\bibitem{wang2021towards}
X.~Wang, X.~Shu, Z.~Zhang, B.~Jiang, Y.~Wang, Y.~Tian, and F.~Wu, ``Towards
  more flexible and accurate object tracking with natural language: Algorithms
  and benchmark,'' in \emph{Proceedings of the IEEE/CVF conference on computer
  vision and pattern recognition}, 2021, pp. 13\,763--13\,773.

\bibitem{guo2024divert}
M.~Guo, Z.~Zhang, L.~Jing, H.~Ling, and H.~Fan, ``Divert more attention to
  vision-language object tracking,'' \emph{IEEE Transactions on Pattern
  Analysis and Machine Intelligence}, 2024.

\bibitem{openai2024gpt4ocard}
\BIBentryALTinterwordspacing
OpenAI, ``Gpt-4o system card,'' 2024. [Online]. Available:
  \url{https://arxiv.org/abs/2410.21276}
\BIBentrySTDinterwordspacing

\bibitem{deepseekai2025deepseekr1incentivizingreasoningcapability}
\BIBentryALTinterwordspacing
DeepSeek-AI, ``Deepseek-r1: Incentivizing reasoning capability in llms via
  reinforcement learning,'' 2025. [Online]. Available:
  \url{https://arxiv.org/abs/2501.12948}
\BIBentrySTDinterwordspacing

\bibitem{bai2023qwentechnicalreport}
\BIBentryALTinterwordspacing
J.~Bai, S.~Bai, and Y.~C. et~al., ``Qwen technical report,'' 2023. [Online].
  Available: \url{https://arxiv.org/abs/2309.16609}
\BIBentrySTDinterwordspacing

\bibitem{yang2025qwen3technicalreport}
\BIBentryALTinterwordspacing
A.~Yang, A.~Li, and B.~Y. et~al., ``Qwen3 technical report,'' 2025. [Online].
  Available: \url{https://arxiv.org/abs/2505.09388}
\BIBentrySTDinterwordspacing

\bibitem{liu2025seg}
Y.~Liu, B.~Peng, Z.~Zhong, Z.~Yue, F.~Lu, B.~Yu, and J.~Jia, ``Seg-zero:
  Reasoning-chain guided segmentation via cognitive reinforcement,''
  \emph{arXiv preprint arXiv:2503.06520}, 2025.

\bibitem{guan2025rstar}
X.~Guan, L.~L. Zhang, Y.~Liu, N.~Shang, Y.~Sun, Y.~Zhu, F.~Yang, and M.~Yang,
  ``rstar-math: Small llms can master math reasoning with self-evolved deep
  thinking,'' \emph{arXiv preprint arXiv:2501.04519}, 2025.

\bibitem{wu2013online}
Y.~Wu, J.~Lim, and M.-H. Yang, ``Online object tracking: A benchmark,'' in
  \emph{Proceedings of the IEEE conference on computer vision and pattern
  recognition}, 2013, pp. 2411--2418.

\bibitem{huang2019got}
L.~Huang, X.~Zhao, and K.~Huang, ``Got-10k: A large high-diversity benchmark
  for generic object tracking in the wild,'' \emph{IEEE transactions on pattern
  analysis and machine intelligence}, vol.~43, no.~5, pp. 1562--1577, 2019.

\bibitem{wang2021tnl2k}
X.~Wang, X.~Shu, Z.~Zhang, B.~Jiang, Y.~Wang, Y.~Tian, and F.~Wu, ``Towards
  more flexible and accurate object tracking with natural language: Algorithms
  and benchmark,'' in \emph{Proceedings of the IEEE/CVF Conference on Computer
  Vision and Pattern Recognition (CVPR)}, June 2021, pp. 13\,763--13\,773.

\bibitem{9339950}
S.~M. Marvasti-Zadeh, L.~Cheng, H.~Ghanei-Yakhdan, and S.~Kasaei, ``Deep
  learning for visual tracking: A comprehensive survey,'' \emph{IEEE
  Transactions on Intelligent Transportation Systems}, vol.~23, no.~5, pp.
  3943--3968, 2022.

\bibitem{hare2016struck}
S.~Hare, S.~Golodetz, A.~Saffari, V.~Vineet, M.-M. Cheng, S.~L. Hicks, and
  P.~H. Torr, ``Struck: Structured output tracking with kernels,'' \emph{IEEE
  transactions on pattern analysis and machine intelligence}, vol.~38, no.~10,
  pp. 2096--2109, 2016.

\bibitem{Nam2015Learning}
H.~Nam and B.~Han, ``Learning multi-domain convolutional neural networks for
  visual tracking,'' in \emph{Proceedings of the IEEE Conference on Computer
  Vision and Pattern Recognition}, 2016, pp. 4293--4302.

\bibitem{SongYiBing_2018_CVPR}
Y.~Song, C.~Ma, X.~Wu, L.~Gong, L.~Bao, W.~Zuo, C.~Shen, R.~W. Lau, and M.-H.
  Yang, ``Vital: Visual tracking via adversarial learning,'' in \emph{The IEEE
  Conference on Computer Vision and Pattern Recognition (CVPR)}, June 2018.

\bibitem{dai2020LTMU}
K.~Dai, Y.~Zhang, D.~Wang, J.~Li, H.~Lu, and X.~Yang, ``High-performance
  long-term tracking with meta-updater,'' in \emph{Proceedings of the IEEE/CVF
  Conference on Computer Vision and Pattern Recognition}, 2020, pp. 6298--6307.

\bibitem{bertinetto2016fully}
L.~Bertinetto, J.~Valmadre, J.~F. Henriques, A.~Vedaldi, and P.~H. Torr,
  ``Fully-convolutional siamese networks for object tracking,'' in
  \emph{Computer vision--ECCV 2016 workshops: Amsterdam, the Netherlands,
  October 8-10 and 15-16, 2016, proceedings, part II 14}.\hskip 1em plus 0.5em
  minus 0.4em\relax Springer, 2016, pp. 850--865.

\bibitem{Tao2016Siamese}
R.~Tao, E.~Gavves, and A.~W. Smeulders, ``Siamese instance search for
  tracking,'' in \emph{Proceedings of the IEEE conference on computer vision
  and pattern recognition}, 2016, pp. 1420--1429.

\bibitem{li2018siamRPN}
B.~Li, J.~Yan, W.~Wu, Z.~Zhu, and X.~Hu, ``High performance visual tracking
  with siamese region proposal network,'' in \emph{Proceedings of the IEEE
  conference on computer vision and pattern recognition}, 2018, pp. 8971--8980.

\bibitem{wang2019SiamMask}
Q.~Wang, L.~Zhang, L.~Bertinetto, W.~Hu, and P.~H. Torr, ``Fast online object
  tracking and segmentation: A unifying approach,'' in \emph{Proceedings of the
  IEEE/CVF conference on Computer Vision and Pattern Recognition}, 2019, pp.
  1328--1338.

\bibitem{xu2020siamfc++}
Y.~Xu, Z.~Wang, Z.~Li, Y.~Yuan, and G.~Yu, ``Siamfc++: Towards robust and
  accurate visual tracking with target estimation guidelines.'' in \emph{AAAI},
  2020, pp. 12\,549--12\,556.

\bibitem{mayer2022ToMP}
C.~Mayer, M.~Danelljan, G.~Bhat, M.~Paul, D.~P. Paudel, F.~Yu, and L.~Van~Gool,
  ``Transforming model prediction for tracking,'' in \emph{Proceedings of the
  IEEE/CVF Conference on Computer Vision and Pattern Recognition}, 2022, pp.
  8731--8740.

\bibitem{chen2021TransT}
X.~Chen, B.~Yan, J.~Zhu, D.~Wang, X.~Yang, and H.~Lu, ``Transformer tracking,''
  in \emph{Proceedings of the IEEE/CVF Conference on Computer Vision and
  Pattern Recognition}, 2021, pp. 8126--8135.

\bibitem{danelljan2015learning}
M.~Danelljan, G.~Hager, F.~Shahbaz~Khan, and M.~Felsberg, ``Learning spatially
  regularized correlation filters for visual tracking,'' in \emph{Proceedings
  of the IEEE International Conference on Computer Vision}, 2015, pp.
  4310--4318.

\bibitem{choi2017attentional}
J.~Choi, H.~J. Chang, S.~Yun, T.~Fischer, Y.~Demiris, J.~Y. Choi \emph{et~al.},
  ``Attentional correlation filter network for adaptive visual tracking.'' in
  \emph{CVPR}, vol.~2, no.~6, 2017, p.~7.

\bibitem{Bolme2010Visual}
D.~S. Bolme, J.~R. Beveridge, B.~A. Draper, and Y.~M. Lui, ``Visual object
  tracking using adaptive correlation filters,'' vol. 119, no.~5, pp.
  2544--2550, 2010.

\bibitem{yao2018joint}
Y.~Yao, X.~Wu, L.~Zhang, S.~Shan, and W.~Zuo, ``Joint representation and
  truncated inference learning for correlation filter based tracking,''
  \emph{arXiv preprint arXiv:1807.11071}, 2018.

\bibitem{valmadre2017end}
J.~Valmadre, L.~Bertinetto, J.~Henriques, A.~Vedaldi, and P.~H. Torr,
  ``End-to-end representation learning for correlation filter based tracking,''
  in \emph{Computer Vision and Pattern Recognition (CVPR), 2017 IEEE Conference
  on}.\hskip 1em plus 0.5em minus 0.4em\relax IEEE, 2017, pp. 5000--5008.

\bibitem{danelljan2019atom}
M.~Danelljan, G.~Bhat, F.~S. Khan, and M.~Felsberg, ``Atom: Accurate tracking
  by overlap maximization,'' in \emph{Proceedings of the IEEE Conference on
  Computer Vision and Pattern Recognition}, 2019, pp. 4660--4669.

\bibitem{bhat2019dimp}
G.~Bhat, M.~Danelljan, L.~V. Gool, and R.~Timofte, ``Learning discriminative
  model prediction for tracking,'' in \emph{Proceedings of the IEEE/CVF
  international conference on computer vision}, 2019, pp. 6182--6191.

\bibitem{danelljan2020prdimp}
M.~Danelljan, L.~V. Gool, and R.~Timofte, ``Probabilistic regression for visual
  tracking,'' in \emph{Proceedings of the IEEE/CVF conference on computer
  vision and pattern recognition}, 2020, pp. 7183--7192.

\bibitem{bhat2020KYS}
G.~Bhat, M.~Danelljan, L.~V. Gool, and R.~Timofte, ``Know your surroundings:
  Exploiting scene information for object tracking,'' in \emph{European
  Conference on Computer Vision}.\hskip 1em plus 0.5em minus 0.4em\relax
  Springer, 2020, pp. 205--221.

\bibitem{2017Attention}
A.~Vaswani, N.~Shazeer, N.~Parmar, J.~Uszkoreit, L.~Jones, A.~N. Gomez,
  L.~Kaiser, and I.~Polosukhin, ``Attention is all you need,'' \emph{arXiv},
  2017.

\bibitem{ye2022ostrack}
B.~Ye, H.~Chang, B.~Ma, S.~Shan, and X.~Chen, ``Joint feature learning and
  relation modeling for tracking: A one-stream framework,'' in \emph{ECCV},
  2022.

\bibitem{sutrack}
X.~Chen, B.~Kang, W.~Geng, J.~Zhu, Y.~Liu, D.~Wang, and H.~Lu, ``Sutrack:
  Towards simple and unified single object tracking,'' 2025.

\bibitem{gao2022aiatrack}
S.~Gao, C.~Zhou, C.~Ma, X.~Wang, and J.~Yuan, ``Aiatrack: Attention in
  attention for transformer visual tracking,'' in \emph{European Conference on
  Computer Vision}.\hskip 1em plus 0.5em minus 0.4em\relax Springer, 2022, pp.
  146--164.

\bibitem{li2025dynamic}
X.~Li, B.~Zhong, Q.~Liang, Z.~Mo, J.~Nong, and S.~Song, ``Dynamic updates for
  language adaptation in visual-language tracking,'' \emph{arXiv preprint
  arXiv:2503.06621}, 2025.

\bibitem{li2017tracking}
Z.~Li, R.~Tao, E.~Gavves, C.~G. Snoek, and A.~W. Smeulders, ``Tracking by
  natural language specification,'' in \emph{Proceedings of the IEEE conference
  on computer vision and pattern recognition}, 2017, pp. 6495--6503.

\bibitem{graves2012long}
A.~Graves and A.~Graves, ``Long short-term memory,'' \emph{Supervised sequence
  labelling with recurrent neural networks}, pp. 37--45, 2012.

\bibitem{simonyan2014very}
K.~Simonyan and A.~Zisserman, ``Very deep convolutional networks for
  large-scale image recognition,'' \emph{arXiv preprint arXiv:1409.1556}, 2014.

\bibitem{wang2018describe}
X.~Wang, C.~Li, R.~Yang, T.~Zhang, J.~Tang, and B.~Luo, ``Describe and attend
  to track: Learning natural language guided structural representation and
  visual attention for object tracking,'' \emph{arXiv preprint
  arXiv:1811.10014}, 2018.

\bibitem{yang2020grounding}
Z.~Yang, T.~Kumar, T.~Chen, J.~Su, and J.~Luo,
  ``Grounding-tracking-integration,'' \emph{IEEE Transactions on Circuits and
  Systems for Video Technology}, vol.~31, no.~9, pp. 3433--3443, 2020.

\bibitem{guo2022divert}
M.~Guo, Z.~Zhang, H.~Fan, and L.~Jing, ``Divert more attention to
  vision-language tracking,'' \emph{Advances in Neural Information Processing
  Systems}, vol.~35, pp. 4446--4460, 2022.

\bibitem{zhang2023all}
C.~Zhang, X.~Sun, Y.~Yang, L.~Liu, Q.~Liu, X.~Zhou, and Y.~Wang, ``All in one:
  Exploring unified vision-language tracking with multi-modal alignment,'' in
  \emph{Proceedings of the 31st ACM International Conference on Multimedia},
  2023, pp. 5552--5561.

\bibitem{zheng2023toward}
Y.~Zheng, B.~Zhong, Q.~Liang, G.~Li, R.~Ji, and X.~Li, ``Toward unified token
  learning for vision-language tracking,'' \emph{IEEE Transactions on Circuits
  and Systems for Video Technology}, vol.~34, no.~4, pp. 2125--2135, 2023.

\bibitem{sun2024chattracker}
Y.~Sun, F.~Yu, S.~Chen, Y.~Zhang, J.~Huang, Y.~Li, C.~Li, and C.~Wang,
  ``Chattracker: Enhancing visual tracking performance via chatting with
  multimodal large language model,'' \emph{Advances in Neural Information
  Processing Systems}, vol.~37, pp. 39\,303--39\,324, 2024.

\bibitem{li2022blip}
J.~Li, D.~Li, C.~Xiong, and S.~Hoi, ``Blip: Bootstrapping language-image
  pre-training for unified vision-language understanding and generation,'' in
  \emph{International conference on machine learning}.\hskip 1em plus 0.5em
  minus 0.4em\relax PMLR, 2022, pp. 12\,888--12\,900.

\bibitem{wei2022chain}
J.~Wei, X.~Wang, D.~Schuurmans, M.~Bosma, F.~Xia, E.~Chi, Q.~V. Le, D.~Zhou
  \emph{et~al.}, ``Chain-of-thought prompting elicits reasoning in large
  language models,'' \emph{Advances in neural information processing systems},
  vol.~35, pp. 24\,824--24\,837, 2022.

\bibitem{kojima2022large}
T.~Kojima, S.~S. Gu, M.~Reid, Y.~Matsuo, and Y.~Iwasawa, ``Large language
  models are zero-shot reasoners,'' \emph{Advances in neural information
  processing systems}, vol.~35, pp. 22\,199--22\,213, 2022.

\bibitem{zhang2024supervised}
X.~Zhang and D.~Ding, ``Supervised chain of thought,'' \emph{arXiv preprint
  arXiv:2410.14198}, 2024.

\bibitem{ouyang2022training}
L.~Ouyang, J.~Wu, X.~Jiang, D.~Almeida, C.~Wainwright, P.~Mishkin, C.~Zhang,
  S.~Agarwal, K.~Slama, A.~Ray \emph{et~al.}, ``Training language models to
  follow instructions with human feedback,'' \emph{Advances in neural
  information processing systems}, vol.~35, pp. 27\,730--27\,744, 2022.

\bibitem{zhang2025r1}
J.~Zhang, J.~Huang, H.~Yao, S.~Liu, X.~Zhang, S.~Lu, and D.~Tao, ``R1-vl:
  Learning to reason with multimodal large language models via step-wise group
  relative policy optimization,'' \emph{arXiv preprint arXiv:2503.12937}, 2025.

\bibitem{wang2025r1}
B.~Wang and W.~Li, ``R1-track: Direct application of mllms to visual object
  tracking via reinforcement learning,'' \emph{arXiv preprint
  arXiv:2506.21980}, 2025.

\bibitem{devlin2019bert}
J.~Devlin, M.-W. Chang, K.~Lee, and K.~Toutanova, ``Bert: Pre-training of deep
  bidirectional transformers for language understanding,'' in \emph{Proceedings
  of the 2019 conference of the North American chapter of the association for
  computational linguistics: human language technologies, volume 1 (long and
  short papers)}, 2019, pp. 4171--4186.

\bibitem{fan2019lasot}
H.~Fan, L.~Lin, F.~Yang, P.~Chu, G.~Deng, S.~Yu, H.~Bai, Y.~Xu, C.~Liao, and
  H.~Ling, ``Lasot: A high-quality benchmark for large-scale single object
  tracking,'' in \emph{Proceedings of the IEEE/CVF conference on computer
  vision and pattern recognition}, 2019, pp. 5374--5383.

\bibitem{Liang2015Encoding}
P.~Liang, E.~Blasch, and H.~Ling, ``Encoding color information for visual
  tracking: Algorithms and benchmark,'' \emph{IEEE transactions on image
  processing}, vol.~24, no.~12, pp. 5630--5644, 2015.

\bibitem{kristan2016novel}
M.~Kristan, J.~Matas, A.~Leonardis, T.~Voj{\'\i}{\v{r}}, R.~Pflugfelder,
  G.~Fernandez, G.~Nebehay, F.~Porikli, and L.~{\v{C}}ehovin, ``A novel
  performance evaluation methodology for single-target trackers,'' \emph{IEEE
  transactions on pattern analysis and machine intelligence}, vol.~38, no.~11,
  pp. 2137--2155, 2016.

\bibitem{li2015nus}
A.~Li, M.~Lin, Y.~Wu, M.-H. Yang, and S.~Yan, ``Nus-pro: A new visual tracking
  challenge,'' \emph{IEEE transactions on pattern analysis and machine
  intelligence}, vol.~38, no.~2, pp. 335--349, 2015.

\bibitem{benchmark2016benchmark}
U.~Benchmark, ``A benchmark and simulator for uav tracking,'' in \emph{European
  conference on computer vision}, vol.~7, 2016.

\bibitem{kiani2017need}
H.~Kiani~Galoogahi, A.~Fagg, C.~Huang, D.~Ramanan, and S.~Lucey, ``Need for
  speed: A benchmark for higher frame rate object tracking,'' in
  \emph{Proceedings of the IEEE international conference on computer vision},
  2017, pp. 1125--1134.

\bibitem{muller2018trackingnet}
M.~Muller, A.~Bibi, S.~Giancola, S.~Alsubaihi, and B.~Ghanem, ``Trackingnet: A
  large-scale dataset and benchmark for object tracking in the wild,'' in
  \emph{Proceedings of the European conference on computer vision (ECCV)},
  2018, pp. 300--317.

\bibitem{valmadre2018long}
J.~Valmadre, L.~Bertinetto, J.~F. Henriques, R.~Tao, A.~Vedaldi, A.~W.
  Smeulders, P.~H. Torr, and E.~Gavves, ``Long-term tracking in the wild: A
  benchmark,'' in \emph{Proceedings of the European conference on computer
  vision (ECCV)}, 2018, pp. 670--685.

\bibitem{10004511}
C.~Zhang, G.~Huang, L.~Liu, S.~Huang, Y.~Yang, X.~Wan, S.~Ge, and D.~Tao,
  ``Webuav-3m: A benchmark for unveiling the power of million-scale deep uav
  tracking,'' \emph{IEEE Transactions on Pattern Analysis and Machine
  Intelligence}, vol.~45, no.~7, pp. 9186--9205, 2023.

\bibitem{zhang2024webuot}
C.~Zhang, L.~Liu, G.~Huang, H.~Wen, X.~Zhou, and Y.~Wang, ``Webuot-1m:
  Advancing deep underwater object tracking with a million-scale benchmark,''
  \emph{Advances in Neural Information Processing Systems}, vol.~37, pp.
  50\,152--50\,167, 2024.

\bibitem{jia2020robust}
S.~Jia, C.~Ma, Y.~Song, and X.~Yang, ``Robust tracking against adversarial
  attacks,'' in \emph{Computer Vision--ECCV 2020: 16th European Conference,
  Glasgow, UK, August 23--28, 2020, Proceedings, Part XIX 16}.\hskip 1em plus
  0.5em minus 0.4em\relax Springer, 2020, pp. 69--84.

\bibitem{cui2022mixformer}
Y.~Cui, C.~Jiang, L.~Wang, and G.~Wu, ``Mixformer: End-to-end tracking with
  iterative mixed attention,'' in \emph{Proceedings of the IEEE/CVF conference
  on computer vision and pattern recognition}, 2022, pp. 13\,608--13\,618.

\bibitem{citetracker}
X.~Li, Y.~Huang, Z.~He, Y.~Wang, H.~Lu, and M.-H. Yang, ``Citetracker:
  Correlating image and text for visual tracking,'' in \emph{ICCV}, 2023.

\bibitem{cai2023robust}
Y.~Cai, J.~Liu, J.~Tang, and G.~Wu, ``Robust object modeling for visual
  tracking,'' in \emph{Proceedings of the IEEE/CVF international conference on
  computer vision}, 2023, pp. 9589--9600.

\bibitem{gao2023generalized}
S.~Gao, C.~Zhou, and J.~Zhang, ``Generalized relation modeling for transformer
  tracking,'' in \emph{Proceedings of the IEEE/CVF conference on computer
  vision and pattern recognition}, 2023, pp. 18\,686--18\,695.

\bibitem{zheng2024odtrack}
Y.~Zheng, B.~Zhong, Q.~Liang, Z.~Mo, S.~Zhang, and X.~Li, ``Odtrack: Online
  dense temporal token learning for visual tracking,'' in \emph{AAAI}, 2024.

\bibitem{shi2024evptrack}
L.~Shi, B.~Zhong, Q.~Liang, N.~Li, S.~Zhang, and X.~Li, ``Explicit visual
  prompts for visual object tracking,'' in \emph{AAAI}, 2024.

\bibitem{xie2024autoregressive}
J.~Xie, B.~Zhong, Z.~Mo, S.~Zhang, L.~Shi, S.~Song, and R.~Ji, ``Autoregressive
  queries for adaptive tracking with spatio-temporal transformers,'' in
  \emph{Proceedings of the IEEE/CVF Conference on Computer Vision and Pattern
  Recognition}, 2024, pp. 19\,300--19\,309.

\bibitem{xu2025less}
C.~Xu, B.~Zhong, Q.~Liang, Y.~Zheng, G.~Li, and S.~Song, ``Less is more: Token
  context-aware learning for object tracking,'' in \emph{Proceedings of the
  AAAI Conference on Artificial Intelligence}, vol.~39, no.~8, 2025, pp.
  8824--8832.

\bibitem{feng2025enhancing}
X.~Feng, D.~Zhang, S.~Hu, X.~Li, M.~Wu, J.~Zhang, X.~Chen, and K.~Huang,
  ``Enhancing vision-language tracking by effectively converting textual cues
  into visual cues,'' in \emph{ICASSP 2025-2025 IEEE International Conference
  on Acoustics, Speech and Signal Processing (ICASSP)}.\hskip 1em plus 0.5em
  minus 0.4em\relax IEEE, 2025, pp. 1--5.

\bibitem{Li2025dutrack}
X.~Li, , B.~Zhong, Q.~Liang, Z.~Mo, J.~Nong, and S.~Song, ``Dynamic updates for
  language adaptation in visual-language tracking,'' in \emph{CVPR}, 2025.

\bibitem{zheng2024llamafactory}
\BIBentryALTinterwordspacing
Y.~Zheng, R.~Zhang, J.~Zhang, Y.~Ye, Z.~Luo, Z.~Feng, and Y.~Ma,
  ``Llamafactory: Unified efficient fine-tuning of 100+ language models,'' in
  \emph{Proceedings of the 62nd Annual Meeting of the Association for
  Computational Linguistics (Volume 3: System Demonstrations)}.\hskip 1em plus
  0.5em minus 0.4em\relax Bangkok, Thailand: Association for Computational
  Linguistics, 2024. [Online]. Available: \url{http://arxiv.org/abs/2403.13372}
\BIBentrySTDinterwordspacing

\bibitem{kwon2023efficient}
W.~Kwon, Z.~Li, S.~Zhuang, Y.~Sheng, L.~Zheng, C.~H. Yu, J.~E. Gonzalez,
  H.~Zhang, and I.~Stoica, ``Efficient memory management for large language
  model serving with pagedattention,'' in \emph{Proceedings of the ACM SIGOPS
  29th Symposium on Operating Systems Principles}, 2023.

\bibitem{zhang2017mdnet}
Z.~Zhang, Y.~Xie, F.~Xing, M.~McGough, and L.~Yang, ``Mdnet: A semantically and
  visually interpretable medical image diagnosis network,'' in
  \emph{Proceedings of the IEEE conference on computer vision and pattern
  recognition}, 2017, pp. 6428--6436.

\bibitem{zhang2020ocean}
Z.~Zhang, H.~Peng, J.~Fu, B.~Li, and W.~Hu, ``Ocean: Object-aware anchor-free
  tracking,'' in \emph{European conference on computer vision}.\hskip 1em plus
  0.5em minus 0.4em\relax Springer, 2020, pp. 771--787.

\bibitem{li2019siamrpn++}
B.~Li, W.~Wu, Q.~Wang, F.~Zhang, J.~Xing, and J.~Yan, ``Siamrpn++: Evolution of
  siamese visual tracking with very deep networks,'' in \emph{Proceedings of
  the IEEE/CVF conference on computer vision and pattern recognition}, 2019,
  pp. 4282--4291.

\bibitem{wang2021transformer}
N.~Wang, W.~Zhou, J.~Wang, and H.~Li, ``Transformer meets tracker: Exploiting
  temporal context for robust visual tracking,'' in \emph{Proceedings of the
  IEEE/CVF conference on computer vision and pattern recognition}, 2021, pp.
  1571--1580.

\bibitem{chen2021transformer}
X.~Chen, B.~Yan, J.~Zhu, D.~Wang, X.~Yang, and H.~Lu, ``Transformer tracking,''
  in \emph{Proceedings of the IEEE/CVF conference on computer vision and
  pattern recognition}, 2021, pp. 8126--8135.

\bibitem{luo2021exploring}
C.~Luo, X.~Yang, and A.~Yuille, ``Exploring simple 3d multi-object tracking for
  autonomous driving,'' in \emph{Proceedings of the IEEE/CVF international
  conference on computer vision}, 2021, pp. 10\,488--10\,497.

\bibitem{xie2023videotrack}
F.~Xie, L.~Chu, J.~Li, Y.~Lu, and C.~Ma, ``Videotrack: Learning to track
  objects via video transformer,'' in \emph{Proceedings of the IEEE/CVF
  conference on computer vision and pattern recognition}, 2023, pp.
  22\,826--22\,835.

\bibitem{chen2023seqtrack}
X.~Chen, H.~Peng, D.~Wang, H.~Lu, and H.~Hu, ``Seqtrack: Sequence to sequence
  learning for visual object tracking,'' in \emph{Proceedings of the IEEE/CVF
  conference on computer vision and pattern recognition}, 2023, pp.
  14\,572--14\,581.

\bibitem{xie2025robust}
J.~Xie, B.~Zhong, Q.~Liang, N.~Li, Z.~Mo, and S.~Song, ``Robust tracking via
  mamba-based context-aware token learning,'' in \emph{Proceedings of the AAAI
  Conference on Artificial Intelligence}, vol.~39, no.~8, 2025, pp. 8727--8735.

\bibitem{li2025mambalct}
X.~Li, B.~Zhong, Q.~Liang, G.~Li, Z.~Mo, and S.~Song, ``Mambalct: Boosting
  tracking via long-term context state space model,'' in \emph{Proceedings of
  the AAAI Conference on Artificial Intelligence}, vol.~39, no.~5, 2025, pp.
  4986--4994.

\bibitem{feng2020real}
Q.~Feng, V.~Ablavsky, Q.~Bai, G.~Li, and S.~Sclaroff, ``Real-time visual object
  tracking with natural language description,'' in \emph{Proceedings of the
  IEEE/CVF winter conference on applications of computer vision}, 2020, pp.
  700--709.

\bibitem{feng2021siamese}
Q.~Feng, V.~Ablavsky, Q.~Bai, and S.~Sclaroff, ``Siamese natural language
  tracker: Tracking by natural language descriptions with siamese trackers,''
  in \emph{Proceedings of the IEEE/CVF conference on computer vision and
  pattern recognition}, 2021, pp. 5851--5860.

\bibitem{zhao2023transformer}
H.~Zhao, X.~Wang, D.~Wang, H.~Lu, and X.~Ruan, ``Transformer vision-language
  tracking via proxy token guided cross-modal fusion,'' \emph{Pattern
  Recognition Letters}, vol. 168, pp. 10--16, 2023.

\bibitem{ge2024consistencies}
J.~Ge, J.~Cao, X.~Zhu, X.~Zhang, C.~Liu, K.~Wang, and B.~Liu, ``Consistencies
  are all you need for semi-supervised vision-language tracking,'' in
  \emph{Proceedings of the 32nd ACM International Conference on Multimedia},
  2024, pp. 1895--1904.

\bibitem{zhang2024one}
G.~Zhang, B.~Zhong, Q.~Liang, Z.~Mo, N.~Li, and S.~Song, ``One-stream stepwise
  decreasing for vision-language tracking,'' \emph{IEEE Transactions on
  Circuits and Systems for Video Technology}, vol.~34, no.~10, pp. 9053--9063,
  2024.

\bibitem{shao2024context}
Y.~Shao, S.~He, Q.~Ye, Y.~Feng, W.~Luo, and J.~Chen, ``Context-aware
  integration of language and visual references for natural language
  tracking,'' in \emph{Proceedings of the IEEE/CVF Conference on Computer
  Vision and Pattern Recognition}, 2024, pp. 19\,208--19\,217.

\bibitem{zhen2025atstrack}
Y.~Zhen, Q.~Wang, Y.~Qiao, L.~Qu, and H.~Fan, ``Atstrack: Enhancing
  visual-language tracking by aligning temporal and spatial scales,''
  \emph{arXiv preprint arXiv:2507.00454}, 2025.

\end{thebibliography}
}

\end{document}